\documentclass[10pt,twocolumn,letterpaper]{article}

\usepackage[pagenumbers]{cvpr} %

\usepackage[dvipsnames]{xcolor}

\definecolor{cvprblue}{rgb}{0.21,0.49,0.74}
\usepackage[pagebackref,breaklinks,colorlinks,citecolor=cvprblue]{hyperref}

\usepackage{times}
\usepackage{epsfig}
\usepackage{graphicx}
\usepackage{amsmath}
\usepackage{amssymb}
\usepackage{comment}

\usepackage{soul}

\usepackage{pifont}
\usepackage{booktabs}
\usepackage{multirow}
\usepackage{threeparttable}
\usepackage{comment}
\usepackage{amsmath,amsfonts,amssymb}
\usepackage{algorithmic}
\usepackage[ruled,vlined]{algorithm2e}
\usepackage{subcaption}
\usepackage{graphicx}
\usepackage{lipsum}
\DeclareMathOperator*{\argmin}{argmin}

\newcommand{\cmark}{\ding{51}}%

\def\paperID{108} %
\def\confName{3DV\xspace}
\def\confYear{2024\xspace}

\begin{document}

\title{
Multi-Body Neural Scene Flow
}

\author{
Kavisha Vidanapathirana\thanks{This work was done during an internship at The Australian Institute for Machine Learning, The University of Adelaide. 
} $^{,1,2}$
\and 
Shin-Fang Chng$^{3}$
\and 
Xueqian Li$^{3}$
\and 
Simon Lucey$^{3}$
\and
$^{1}$ \text{Queensland University of Technology} %
\and 
$^{2}$ \text{CSIRO Robotics}%
\and 
$^{3}$ \text{The University of Adelaide}
}

\maketitle

\begin{abstract}
The test-time optimization of scene flow---using a coordinate network as a neural prior~\cite{li2021nsfp}---has gained popularity due to its simplicity, lack of dataset bias, and state-of-the-art performance. We observe, however, that although coordinate networks capture general motions by implicitly regularizing the scene flow predictions to be spatially smooth, the neural prior by itself is unable to identify the underlying multi-body rigid motions present in real-world data.
To address this, 
we show that multi-body rigidity can be achieved without the cumbersome and brittle strategy of constraining the $SE(3)$ parameters of each rigid body as done in previous works. This is achieved by regularizing the scene flow optimization to encourage isometry in flow predictions for rigid bodies. 
This strategy enables multi-body rigidity in scene flow while maintaining a continuous flow field, 
hence allowing dense long-term scene flow integration across a sequence of point clouds. 
We conduct extensive experiments on real-world datasets 
and demonstrate that our approach outperforms the state-of-the-art in 3D scene flow and long-term point-wise 4D trajectory prediction.
The code is available at: \href{https://github.com/kavisha725/MBNSF}{https://github.com/kavisha725/MBNSF}.
\end{abstract}    
\section{Introduction}
\label{sec:intro}

Scene flow is a crucial task in dynamic 3D perception.
A dense scene flow yields a high-fidelity representation of the dynamic environment, which enables various tasks such as ego-motion estimation~\cite{deng2023rsf}, motion segmentation~\cite{baur2021slim}, object detection~\cite{najibi2022motion}, dynamic object reconstruction~\cite{huang2022dynamic}, and point cloud densification~\cite{wang2022ntp}.
An emerging paradigm for estimating scene flow is to utilize
a coordinate MLP to represent the motion field between two point clouds~\cite{li2021nsfp}. 
The so-called~\textit{Neural Scene Flow Prior} (NSFP)~\cite{li2021nsfp} is particularly appealing as it does not require training data (labeled or unlabeled) yet consistently outperforms existing learning methods in open-world conditions. NSFP implicitly enforces the motion field to be~\textit{spatially smooth} via the regularization properties of the MLP architecture.

\begin{figure}[t!]
\centering
\includegraphics[width=0.9\linewidth]{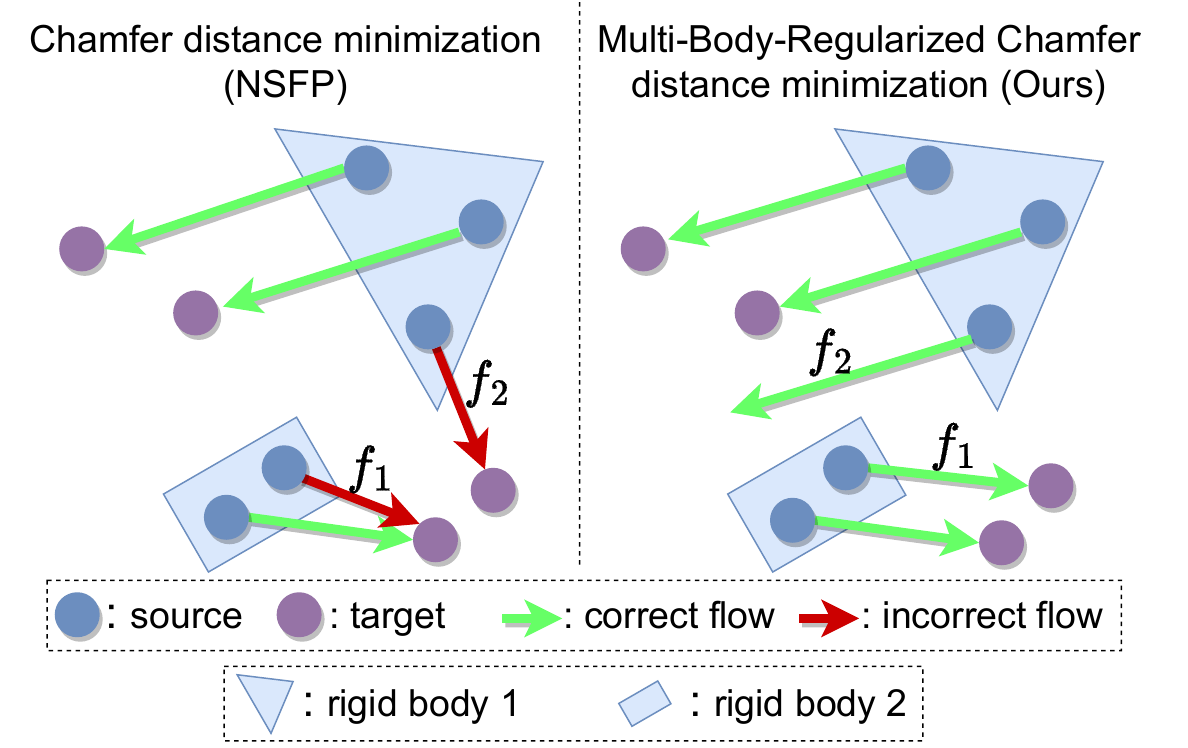}
\caption{2D visualization of 3D flow predictions for points sampled from two rigid bodies. 
Relying solely on Chamfer distance minimization, NSFP (left)
may violate multi-body rigidity due to either (1) the flow predictions for different points collapsing to a single target ($ \mathbf{f_1} $), or (2) a point being assigned a target in a different body ($ \mathbf{f_2} $) --- which happens to be closer due to the inconsistent sampling of the scene by LiDAR sensors. 
Our approach (right) encourages multi-body-rigidity and enables predicting accurate motion vectors even for points that don't have a valid target ($ \mathbf{f_2} $).}
\label{fig:front_page}
\end{figure}

While spatial smoothness addresses general motions in natural environments, it does not account for the independent motions of multiple rigid objects. 
For example,
the flow field of two nearby cars moving in opposite directions may not be smooth near the boundaries. 
Inevitably, MLP networks optimized without rigid motion constraints may predict  physically implausible motion vectors.
A contrived example of this is shown in~\cref{fig:front_page}, where the potential failure cases of unconstrained optimization can easily be avoided via our proposed multi-body regularization. 

To enforce multi-body rigidity, prior work has leveraged geometric priors of the environment~\cite{gojcic2021weakly, huang2022dynamic, najibi2022motion, dong2022exploiting, li2022rigidflow}. Specifically, these methods assume that autonomous driving scenarios consist of a static background and a foreground composed of rigidly moving entities, which can be approximated using spatial clustering. %
Despite addressing common failure cases, \eg, predicting physically implausible motion vectors, these methods have limitations. Notably, they enforce rigidity by either (1) using regularizers that do not enforce full $SE(3)$ rigidity~\cite{pontes2020graphlaplacian, kittenplon2021flowstep3d, najibi2022motion}(\ie, not constraining both the translation and rotation components of motion), or (2) enforce rigidity via the cumbersome estimation of $SE(3)$ parameters of each rigid body~\cite{gojcic2021weakly, dong2022exploiting, li2022rigidflow, huang2022dynamic, deng2023rsf}. Additionally, these methods are dependent on accurate estimation of rigid bodies, and output \textit{discrete} flow estimates as opposed to continuous motion fields.

In contrast, we take a fundamentally distinct approach to enforcing multi-body rigidity. We regularize the flow predictions of rigid bodies to be an \textit{approximate isometry} by constraining the relative distances of their points to remain almost constant. This enforces $SE(3)$ rigidity without explicitly constraining the $SE(3)$ parameters of rigid bodies. This approach is not dependent on accurate rigid body extraction and maintains a continuous motion field allowing long-term trajectory estimation via (1) scene flow integration or (2) use in tandem with temporal regularizers \cite{wang2022ntp}.

In summary, we introduce \textit{Multi-Body Neural Scene Flow} (MBNSF) and show that regularizing the optimization of neural scene flow --- to accommodate the independent motions of multiple rigid bodies --- can prevent the prediction of physically implausible motion vectors. As our main contributions, we demonstrate that: 
\begin{itemize}
    \item Multi-body rigidity can be achieved without constraining the $SE(3)$ parameters of each rigid body. This maintains a continuous motion field, allowing accurate flow integration beyond a single pair of point clouds.
    \item This strategy is not dependent on the exact estimation of rigid bodies present in the scene --- which is a difficult task given the sparsity of LiDAR data.
    \item The generalizability of this approach can be seen by extending it to the task of direct 4D trajectory estimation. This is shown by the introduction of \textit{Multi-Body Neural Trajectories} (MBNT) via integration with NTP~\cite{wang2022ntp}.
\end{itemize} 

Our experimental results demonstrate how our approach enforces multi-body rigidity and outperforms the state-of-the-art in scene flow and long-term trajectory prediction.

\section{Related Work}
\label{sec:related}
\subsection{Scene Flow and Continuous Motion Fields}
The scene flow problem was first addressed by Vedula~\etal~\cite{vedula1999threedimsceneflow} using stereo RGB data. Recent works mainly focus on 3D LiDAR data acquired in autonomous driving settings, which poses additional challenges due to the representation of highly dynamic and large outdoor scenes using point clouds with sampling sparsity and large numbers of points. Broadly, the current state-of-the-art can be divided into two categories: (1) data-driven methods that utilize full~\cite{liu2019flownet3d, behl2019pointflownet, wang2020flownet3d++, li2021hcrf, wang2021hierarchical, liu2019meteornet, gu2019hplflownet, puy2020flot, wang2021festa, wei2021pv, cheng2022bi, battrawy2022rms, wang2022matters, wang20233d, fu2023pt, wu2023pointconvformer}/weak~\cite{gojcic2021weakly, dong2022exploiting, huang2022dynamic}/self~\cite{mittal2020just, tishchenko2020self, wu2020pointpwc, baur2021slim, kittenplon2021flowstep3d, lang2023scoop}-supervised learning on large datasets to provide efficient inference on data within the same distribution, and (2) test-time optimization methods~\cite{pontes2020graphlaplacian, li2021nsfp} that optimize scene flow predictions independently for each test sample, hence trading off efficiency for better generalization and independence from training data. 

The majority of scene flow methods predict discrete flow vectors for each point in the source point cloud. A separate line of work aims to estimate a continuous motion field between the two point clouds from which discrete motion vectors can be sampled~\cite{li2021nsfp}. The continuous formulation enables the direct estimation of dense long-term point trajectories across a sequence of point clouds, and the estimation of flow for points not present in the initial sparse point clouds. 
We aim to incorporate multi-body rigidity while still maintaining this continuous representation. 

\subsection{Multi-Body Rigidity}
Methods that enforce multi-body rigidity have become ubiquitous in scene flow literature. 
While the initial works of Costeira and Kanade~\cite{costeira1998multibody}, Vogel~\etal~\cite{vogel20113d, vogel2013piecewise, vogel20153d}, and more recent works~\cite{quiroga2014dense, menze2015object, ma2019deep, huang2019clusterslam, teed2021raft3d} have explored multi-body rigidity using stereo or RGBD data, we focus our discussion on methods that operate on LiDAR point clouds. 

Dewan~\etal~\cite{dewan2016rigid} first explored rigid flow on LiDAR data using spatial smoothing on a factor graph. PointFlowNet~\cite{behl2019pointflownet} utilized object detection to enable multi-body rigidity but only considered cars as moving objects. Pontes~\etal~\cite{pontes2020graphlaplacian} proposed graph-Laplacian regularization for ``as-rigid-as-possible'' scene flow without explicitly enforcing multi-body rigidity; thus is not capable of identifying separate objects or enforcing sharp boundaries between them. Subsequent methods have all incorporated some form of scene decomposition/clustering in order to identify the rigid objects explicitly. 
Our method takes this approach.

WsRSF~\cite{gojcic2021weakly} utilized point cloud clustering using DBSCAN~\cite{ester1996dbscan} to enforce rigidity via estimating $SE(3)$ parameters per cluster.
Subsequent works such as~\cite{dong2022exploiting, chodosh2023re, deng2023rsf} have also utilized DBSCAN for rigid body extraction.
More recently, RSF~\cite{deng2023rsf} proposed 
using differentiable bounding boxes to represent rigid bodies. %
However, such an approach is limited to settings where objects can be parameterized using boxes. Hence, our approach also utilizes DBSCAN clustering for rigid body extraction. 
 Unlike previous methods, we demonstrate that our method is not dependent on the exact estimation of rigid bodies into one cluster\footnote{Further comparison of these methods with ours is provided in~\cref{sec:sup_related_work}.}.
\vspace{-1em}
\paragraph{Rigidity in Neural Scene Flow}
This paper focuses on enforcing multi-body rigidity on scene flow optimized using neural priors, specifically NSFP~\cite{li2021nsfp}. 
Najibi~\etal~\cite{najibi2022motion} proposed NSFP++,
which optimizes an NSFP network on each cluster extracted from DBSCAN while additionally using a ``flow consistency'' regularizer to constrain the flow vectors within a cluster to be equal. 
We note that, due to only constraining the translation component of flow predictions, NSFP++ is still susceptible to physically implausible predictions as the rotation is unconstrained. 

Current methods which enforce the full $SE(3)$ rigidity do so by explicitly estimating and constraining the $SE(3)$ parameters of each rigid body~\cite{gojcic2021weakly, dong2022exploiting, li2022rigidflow, huang2022dynamic, chodosh2023re, deng2023rsf}. 
Recently, Chodosh~\etal~\cite{chodosh2023re} proposed to post-process NSFP predictions to obtain multi-body rigidity. Their formulation requires explicitly constraining the SE(3) parameters of each rigid body, and violates the continuous representation of NSFP.
We circumvent this cumbersome strategy and pioneer the enforcement of multi-body rigidity in neural scene flow while maintaining a continuous flow field.

\section{Method}
\label{sec:method}

\subsection{Problem Formulation}
\label{sec:method_problem}

\paragraph{Continuous Motion Fields} 
Given a temporal sequence of $T$ point clouds represented as $ \{\mathbf{P}_i\}_{i=1}^{T}$, where each point cloud $ \mathbf{P}_i \:{\in}\: \mathbb{R}^{N_i {\times} 3}  $ contains a varying number of $N_i$ 3D points, the goal is to estimate the motion of each point between any given timestamps within the interval $[1,T]$. The solution to this problem is the 4D motion field of points within this time interval and can be represented as a function $\Phi(\cdot; \mathbf{\theta})$, parameterized by $\theta$ such that
\begin{equation}
    \mathbf{f} = \Phi(\mathbf{p},t,\hat{t}; \mathbf{\theta}), %
    \label{eq:motion_field}
\end{equation}
where for a point $\mathbf{p}$ at time $t$, $\mathbf{\hat{p} \:{=}\: p \:{+}\: f}$ is its position at time $\hat{t}$ ($\mathbf{p,\hat{p},f} \:{\in}\: \mathbb{R}^3$). The motion field above is said to be \textit{continuous} if the output of $\Phi$ varies continuously with respect to continuous variation in its input arguments. 
\vspace{-1em}
\paragraph{Scene Flow}
Scene flow is a simplification of the above to a single pair of consecutive point clouds. Let $ \mathbf{P}_{1} \:{\in}\: \mathbb{R}^{N_1 {\times} 3}$ and $\mathbf{P}_{2} \:{\in}\: \mathbb{R}^{N_{2} {\times} 3} $ be two point clouds sampled at time $t_1$ and $t_2$, respectively. Scene flow models the motion of each point in $\mathbf{P}_{1}$ from time $t_1$ to $t_2$ as 
\begin{equation}
    \mathbf{F}_{1} =  \Phi(\mathbf{P}_{1}; \theta) \in \mathbb{R}^{N_1 \times 3}.
    \label{eq:3d_scene_flow}
\end{equation}
Note that due to the inconsistent sampling of points in LiDAR point clouds, the number of points in the point clouds $\mathbf{P}_1$ and $\mathbf{P}_2$ are typically different and there may not be an exact correspondence between them.

For brevity, we explain our method in the context of scene flow. Note that our method is versatile and can be easily extended for long-term trajectories using the formulation in~\cref{eq:motion_field}, as demonstrated in our experiments in~\cref{sec:traj_eval}.

\vspace{-1em}
\paragraph{Optimization}
Following NSFP~\cite{li2021nsfp}, we optimize the continuous motion field as
\begin{align}
    \mathbf{\theta}^{*} & = \argmin_{\mathbf{\theta}} \,  \mathcal{L}_{CD}\Bigl(\mathbf{P}_{1} + \Phi(\mathbf{P}_{1}; \theta), \mathbf{P}_{2}\Bigr),%
\label{eq:nsfp_opt}
\end{align}
which minimizes the truncated Chamfer Distance ($\mathcal{L}_{CD}$) between the projection of $\mathbf{P}_{1}$ (using the predicted flow) and $\mathbf{P}_{2}$, using the smoothness prior of an MLP coordinate network $\Phi$. While this optimization scheme produces a continuous and smooth flow field estimation, for accurate predictions in environments with multiple independent dynamic objects, we require additional regularization to (1) enable sharp boundaries between different objects, and to (2) ensure that the predicted flow is rigid within each object.

\subsection{Multi-Body Regularization}
\label{sec:method_spatial_reg}

In this section, we discuss the relationship between isometry and $SE(3)$ rigidity, and describe our method for regularizing multi-body rigidity in neural scene flow estimation (summarized in~\cref{alg:method}).
\vspace{-1em}
\paragraph{Isometric Flow}
Our goal is to ensure $SE(3)$ rigidity in the flow predictions of rigid bodies.
As shown by Beckman and Quarles~\cite{beckman1953isometries}, a mapping in Euclidean space that preserves the pairwise distances between elements (\ie, an isometry) implies a rigid Euclidean transformation 
(\ie, $E(3)$ for the case of 3D space).  
Extending this theorem to the task of scene flow yields the following corollary:

\noindent \textbf{Corollary} (Rigid flow via isometry): \textit{Constraining  the relative distances between points of a rigid body to remain unchanged after the flow is equivalent to applying a rigid transformation to that body. 
Formally,}
\begin{equation}
    \left\Vert \mathbf{p}_i-\mathbf{p}_j\right\Vert \:{=}\: \left\Vert \mathbf{\hat{p}}_i-\mathbf{\hat{p}}_j\right\Vert \; \forall i,j \:{\implies}\: \mathbf{\hat{p}}_i \:{=}\: \mathbf{T} \circ \mathbf{p}_i \;\; \forall i,
\label{eq:isometric_flow}
\end{equation}
\textit{where $\mathbf{\hat{p}}$ is the position of $\mathbf{p}$ after flow, and $\mathbf{T} \:{\circ}\: \mathbf{p}$ represents applying the transformation $\mathbf{T} \:{\in}\: E(3)$ to the point $\mathbf{p}$.
Additionally, as the scene-level spatial smoothness property of the neural prior will discourage flow predictions that result in object-level reflections, the transformation $\mathbf{T} \:{\in}\: E(3)$ is (in practice) limited to the subgroup of special Euclidean transformations, \ie, $SE(3)$. }

Hence, instead of explicitly estimating an $SE(3)$ transformation for each rigid body, we can obtain $SE(3)$ rigidity by ensuring that the flow prediction of each body maintains an isometry. 
Ideally, the flow for a rigid body should maintain a strict isometry. However, enforcing strict isometry is often impractical due to sensor noise---so we instead aim for an approximate isometry~\footnote{Approximate isometry is an isometry that allows a margin of error~\cite{bhatia1997approximate}.}.%
\vspace{-1em}
\paragraph{Rigid Body Extraction} We note that isometry should be preserved independently for each rigid body in the scene. We extract rigid bodies by decomposing the point cloud into spatial clusters using $\mathtt{DBSCAN}$~\cite{ester1996dbscan}, inline with previous works on multi-body rigidity~\cite{gojcic2021weakly, dong2022exploiting, huang2022dynamic, deng2023rsf, najibi2022motion}. Thus, the source point cloud $ \mathbf{P}_1 $ is represented by a set of clusters
$ \{\mathbf{C}_1, \dots, \mathbf{C}_m\} $, where each cluster
is a set of 3D points. The number of clusters $ m $ varies depending on the environment, the LiDAR sensor, and the $\mathtt{DBSCAN}$ parameters. 

As noted in~\cref{sec:related}, clustering methods such as $\mathtt{DBSCAN}$ are a heuristic for identifying rigid bodies in a point cloud and often contain errors. One way to handle this is by over-clustering (\ie, more clusters than rigid bodies) the point cloud. This strategy 
comes at the cost of splitting single rigid bodies into multiple clusters. 
We explore a new approach to multi-body rigidity 
such that over-clustering does not hamper scene flow performance (see~\cref{fig:dbscan_param}).

\vspace{-1em}
\paragraph{Pairwise Distance Constraint}
Given the clusters $ \{\mathbf{C}_1, \dots, \mathbf{C}_m\} $ in $\mathbf{P}_1$, we aim to incorporate an approximate isometry on the flow of each cluster to regularise the optimization in~\cref{eq:nsfp_opt}. We define an undirected graph $ \mathcal{G} \:{=}\:  (\mathcal{V},\mathcal{E}) $ between each cluster $ \mathbf{C}$~\footnote{For brevity, we drop the subscript cluster index.} and its projection $ \mathbf{\hat{C}} \:{=}\:  \mathbf{C} \:{+}\: \mathbf{F} $.
The vertices $ \mathcal{V} \:{=}\: \{1, .., n\} $ represent $\mathbf{F}$, where $\mathbf{F} \:{=}\: \{\mathbf{f}_1,\dots,\mathbf{f}_n\}$ represents the flow of the $ n $ points in $ \mathbf{C}$.
The edges of the graph $ \mathcal{E} $ represent the degree of preservation of the pair-wise distances between points in $ \mathbf{C}$ after projecting them using  $\mathbf{F}$.
Thus, the adjacency matrix of this graph is a function of both $ \mathbf{C}$ and $\mathbf{F}$; and is represented as $\mathbf{A}(\mathbf{C}, \mathbf{F}) \:{\in}\: \mathbb{R}^{n {\times} n}$ (noted as $\mathbf{A}$ for brevity). 
Specifically, each element $ \mathbf{A}[i,j] $ in $\mathbf{A}$ 
is a score representing the degree of preservation of isometry 
between the flows $\mathbf{f}_i, \mathbf{f}_j$. 

While there are multiple options for defining such a score, for enforcing approximate isometry, we require a function that gives the maximum score for exact relative distance preservation (strict isometry) while smoothly decreasing to zero for pairs that diverge more than a select threshold (approximate isometry). 
To this end, we adopt the scoring of Bai~\etal~\cite{bai2021pointdsc} (for correspondence filtering) to the task of scene flow as
\begin{equation}
    \mathbf{A}[i,j] = \left[1 - \frac{(d_{i,j} - \hat{d}_{i,j})^{2}}{d_{thr}^{2}}\right]_+,
\label{eq:spatial_consistency}    
\end{equation}
where
\begin{equation}
    d_{i,j} = \left\Vert \mathbf{p}_i-\mathbf{p}_j\right\Vert  ,  \quad \mathbf{p} \in \mathbf{C},
\end{equation}
and 
\begin{equation}
    \hat{d}_{i,j} = \left\Vert \mathbf{\hat{p}}_i-\mathbf{\hat{p}}_j\right\Vert,  \quad \mathbf{\hat{p} = (p + f)} \in \mathbf{\hat{C}}.
\end{equation}

\begin{figure}[t]
\centering
\includegraphics[width=0.99\linewidth]{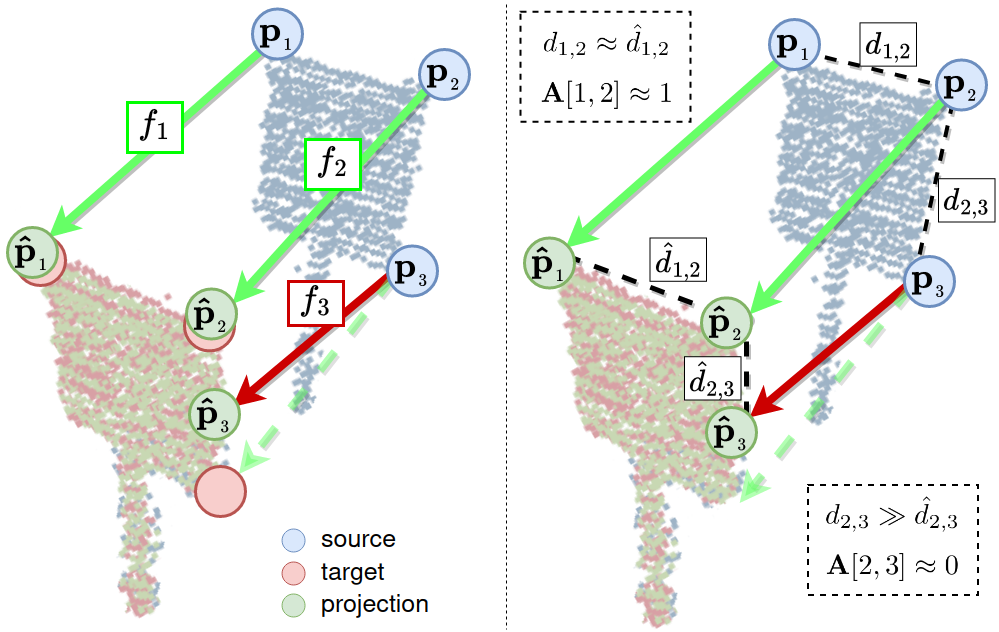}
\caption{Enforcing approximate-isometry in scene flow of rigid bodies. The source point cluster (blue) representing a rigid body (a signpost in this example) is projected (green) to closely match its target (red). Out of the three depicted flow vectors, $\mathbf{f}_1$ and $\mathbf{f}_2$ maintain an approximate-isometry (\ie $d_{1,2} \approx \hat{d}_{1,2} $), and thus get a high score in $\mathbf{A}$ (\ie high $\mathbf{A}[1,2]$). $\mathbf{f}_3$ is a noisy flow prediction that violates the approximate-isometry and gets low scores in $\mathbf{A}$ in relation to other flows (\ie low $\mathbf{A}[i,3], \forall i \neq 3$).}
\label{fig:method}
\end{figure}

$\mathbf{A}[i,j]$ measures how well the relative distance of two points ($\mathbf{p}_i, \mathbf{p}_j \:{\in}\: \mathbf{C}$) is preserved after projecting with their predicted flow ($\mathbf{f}_i, \mathbf{f}_j$) to obtain $\mathbf{\hat{p}}_i, \mathbf{\hat{p}}_j \:{\in}\: \mathbf{\hat{C}}$.
The parameter $d_{thr}$ determines the level of sensitivity to the difference in relative length~\footnote{We include analysis on the sensitivity of accuracy and efficiency of our method to $d_{thr}$ in Sec. \ref{sec:sup_dthr_sensitivity}.} and $[\cdot]_+\:{=}\:\max (\cdot, 0)$. 
If the difference in relative distance between a pair of flows exceeds $d_{thr}$, the corresponding score in $\mathbf{A}$ will be $0$. Conversely, if the relative distance is preserved exactly, the score will be $1$; see~\cref{fig:method} for an example.

Given this adjacency matrix, a non-negative score %
representing the degree of the preservation of isometry within the cluster $\mathbf{C}$
can be computed as
 \begin{equation}
     s(\mathbf{C}, \mathbf{F}) = \frac{1}{n^2} \,\mathbf{1^{\intercal}A1},
 \label{eq:sc_score}
 \end{equation}
where $\mathbf{1} \:{\in}\: \mathbb{R}^{n {\times} 1} $ is a column vector of all ones. This represents the average of all elements in $\mathbf{A}$. 
Maximizing this score involves enforcing \emph{all} flow pairs of a cluster to maintain their pairwise relationships. 
However, we observe that this optimization objective can be overly strict as point clouds obtained from LiDAR sensors typically contain noisy points. 
To address this, we formulate a softer learning objective to optimize the pairwise constraints of a rigid body, which is robust to this noise. 
\vspace{-1em}
\paragraph{Robust Pairwise Distance Constraint}
Towards this end, inspired by the work of Leordeanu and Hebert~\cite{leordeanu2005spectral}, we reformulate~\cref{eq:sc_score} as
\begin{equation}
     s(\mathbf{C}, \mathbf{F}) = \frac{1}{n^2} \max_{\mathbf{v} \in \mathbb{B}^{n}}\,\mathbf{v^{\intercal}Av},
 \label{eq:sc_score_new}
 \end{equation}
where column vectors ($\mathbf{1}$) in~\cref{eq:sc_score} are replaced with a binary indicator vector $\mathbf{v} \:{\in}\: \mathbb{B}^{n}$. 
The role of vector $\mathbf{v}$ is to select flow predictions corresponding to noise-free points such that $\mathbf{v}[i] \:{=}\: 1$ if $\mathbf{f}_i$ is noise-free and $0$ otherwise. 
Our motivation for this reformulation is that the graph $ \mathcal{G} $ forms clusters of nodes (flow predictions) that have high consistency with each other, and the main cluster of this graph is likely to consist of nodes that are free of noise---as it is  unlikely that the flows of noisy points will maintain consistency with those of noise-free points, or within themselves. 
Thus, the maximization of~\cref{eq:sc_score_new} requires the indicator vector representing the main cluster in $ \mathcal{G} $.

Following the relaxations utilized in~\cite{leordeanu2005spectral}, by relaxing the integer constraint on $\mathbf{v}$ and fixing its norm to be $1$, the vector $\mathbf{v}$ that maximizes~\cref{eq:sc_score_new} can be approximated using the leading eigenvector of $\mathbf{A}$ according to the Rayleigh quotient. Thus, we obtain the maximized score as
\begin{equation}
     s(\mathbf{C}, \mathbf{F}) = \frac{1}{n} \,\mathbf{v^{*\intercal}Av^{*}},%
 \label{eq:sc_score_final}
 \end{equation}
 where $\mathbf{v}^{*}$ is the leading eigenvector of $\mathbf{A}$, and the normalizing factor ($n^2$) in~\cref{eq:sc_score_new} is reduced to $n$ as $\mathbf{v}^*$ is now normalized. 
 In practice, the leading eigenvector can be efficiently estimated using the power-iteration method.
 The score of the entire scene can be obtained using the average score of all clusters in the source point cloud as
\begin{equation}
  s_{avg}(\mathbf{P}_{1}; \theta) =  \frac{1}{m}  \sum_{i=1}^{m} s(\mathbf{C}_{i}, \mathbf{F}_{i}), %
\end{equation}
where~$\mathbf{F}_{i}$ is the flow of cluster~$\mathbf{C}_{i}$ and~$\theta$ denotes the parameters of the neural network describing the entire flow field. 

\vspace{-1em}

\paragraph{Isometric Flow Regularization}
Using this robust score, we introduce a multi-body regularization term (${L}_{MB}$) as
\begin{equation}
  \mathcal{L}_{MB}\bigl(\mathbf{P}_{1}; \mathbf{\theta}\bigr) = -\log \bigl(  s_{avg}(\mathbf{P}_{1}; \theta) \bigr),
  \label{eq:neg_log}
\end{equation}
which encourages approximate isometry for the flow of all clusters. 
We select the negative log function due to its monotonicity, convexity, and exponential penalization of low values of $s(\mathbf{C}, \mathbf{F})$. 
By maximizing $s(\mathbf{C}, \mathbf{F})$ for each cluster, the resulting flow predictions of that cluster will be an approximate isometry preserving the relative distances between points after projecting with the predicted flow, and hence represent a rigid motion for each cluster according to~\cref{eq:isometric_flow}. 
\cref{alg:method} summarizes this calculation.

The optimization objective in~\cref{eq:nsfp_opt} is extended as:
\begin{align}
    \mathbf{\theta}^{*} = \argmin_{\mathbf{\theta}} \,  \mathcal{L}_{CD}\Bigl(\mathbf{P}_{1}  + \Phi(\mathbf{P}_{1}; \theta), \mathbf{P}_{2} \Bigr) \notag \\  + \quad \omega \cdot \mathcal{L}_{MB}\Bigl(\mathbf{P}_{1}; \mathbf{\theta} \Bigr),
    \label{eq:final_objective}
\end{align}
where $\omega$ is a scalar weight~\footnote{We include analysis on the sensitivity of accuracy and efficiency of our method to $\omega$ in~\cref{sec:sup_omega_sensitivity}.}. $\mathcal{L}_{CD}$ encourages deforming the source point cloud $\mathbf{P}_1$ in order to match the target $\mathbf{P}_2$, and $\mathcal{L}_{MB}$ enforces the preservation of relative point distances within all rigid bodies of $\mathbf{P}_1$ --- resulting in multi-body rigidity. 
Unlike existing methods which require the explicit estimation of $SE(3)$ parameters for each rigid body~\cite{gojcic2021weakly, dong2022exploiting, li2022rigidflow, huang2022dynamic, deng2023rsf}, this formulation enforces multi-body $SE(3)$ rigidity without such explicit estimation. As a result, our method estimates continuous motion fields that enable tasks such as flow interpolation and long-term flow integration.
Details on the integration of our regularizer with NSFP \cite{li2021nsfp} for 3D scene flow prediction, and NTP \cite{wang2022ntp} for 4D trajectory prediction are provided in Sec. \ref{sec:sup_integration_details}.

\SetKwComment{Comment}{\# }{.}

\begin{algorithm}[t]
\SetAlgoLined
\caption{Multi-Body Regularizer}\label{alg:method}
\KwData{  $ \mathbf{P}, \mathbf{F}$ \Comment*[r]{Source cloud, flow} 
        } 
\KwResult{$\mathcal{L}_{MB}$\Comment*[r]{Multi-body loss} }
$\mathbf{\hat{P}} \gets \mathbf{P} + \mathbf{F}$\;
$\{\mathbf{I}_i\} \gets \mathtt{DBSCAN}(\mathbf{P})$ \Comment*[r]{cluster indices} 
$\{\mathbf{C}_i\} \gets \mathtt{ExtractClusters}(\mathbf{P}, \{\mathbf{I}_i\})$\;
$\{\mathbf{\hat{C}}_i\} \gets \mathtt{ExtractClusters}(\mathbf{\hat{P}}, \{\mathbf{I}_i\})$\;
$s_{sum} \gets  0 $\;
\For{$ \mathbf{C}_i \in \{\mathbf{C}_i\} $}{
    $\mathcal{G} \gets \mathtt{FormulateGraph}(\mathbf{C}_i , \mathbf{\hat{C}}_i)$\;
    $\mathbf{A} \gets \mathtt{AdjacencyMatrix}(\mathcal{G})$\Comment*[r]{\cref{eq:spatial_consistency}}
    $\mathbf{v}^* \gets \mathtt{PowerIteration}(\mathbf{A})$\;
    $s \gets \frac{1}{|\mathbf{C}_i|} \,\mathbf{v^{*\intercal}Av^{*}} $\Comment*[r]{\cref{eq:sc_score_final}}
    $s_{sum} \gets  s_{sum} + s $\;
}
$s_{avg} \gets  \frac{1}{|\{\mathbf{C}_i\}|} s_{sum} $\;
$\mathcal{L}_{MB} \gets -\log(s_{avg})$\Comment*[r]{\cref{eq:neg_log}}
\end{algorithm}

\section{Experimental Evaluation}
\label{sec:results}

\subsection{Evaluation Setup}

\paragraph{Scene Flow Baselines} For learning-based methods, we test the fully-supervised FLOT~\cite{puy2020flot} and the self-supervised PointPWC-Net~\cite{wu2020pointpwc}, which are trained on the synthetic FlyingThings3D~\cite{mayer2016large} dataset. We also test WsRSF~\cite{gojcic2021weakly}, which enforces multi-body rigidity in a weakly-supervised manner. For learning-based methods, we use the official implementations and pre-trained models (without fine-tuning). 
For unsupervised (test-time optimization) methods, we test NSFP~\cite{li2021nsfp} using the original implementation, and we implement NSFP++~\cite{najibi2022motion} (noted as NSFP++$\dagger$) as its implementation is not publicly available~\footnote{\label{fn:code}We commit to open-sourcing our in-house implementations of NSFP++$\dagger$, NSFP$\dagger$, NTP$\dagger$, MBNSF and MBNT.}. 
We incorporate our regularizer with NSFP (noted as MBNSF) using identical hyper-parameters: a learning rate of 0.003, and a maximum of 1000 iterations with early-stopping patience set to 100. 

\vspace{-1em}
\paragraph{Trajectory Estimation Baselines} 
Predicting continuous motion fields enables the estimation of dense long-term trajectories using forward-Euler~\cite{li2021nsfp} integration. 
Additionally, the generalizability of our approach enables integration with direct 4D trajectory prediction methods~\cite{wang2022ntp}. 
We implement forward-Euler integration with NSFP (noted as NSFP$\dagger$) and also implement NTP~\cite{wang2022ntp} (noted as NTP$\dagger$).
We integrate our regularizer to the above under identical hyper-parameters noted as MBNSF and MBNT, respectively\footref{fn:code}.

\vspace{-1em}
\paragraph{Datasets}
We focus our evaluation on 
real-world, large-scale, long-range LiDAR datasets in dynamic urban environments pertinent to autonomous driving applications. 
We select the two recent datasets: Argoverse~\cite{chang2019argoverse} and Waymo Open~\cite{sun2020scalabilitywaymodataset}. Since ground-truth scene flow annotations are not provided in the original release of these datasets, we generate ground-truth following the strategies in~\cite{pontes2020graphlaplacian, li2021nsfp, wang2022ntp} for Argoverse and~\cite{jin2022deformation, jund2021scalable} for Waymo. 

\begin{table*}[!t]
    \caption{Scene flow evaluation on Waymo and Argoverse datasets. Learning-based methods that only support 8192 points are evaluated separately and drastically underperform in out-of-distribution settings. NSFP\cite{li2021nsfp} and MBNSF (Ours) predict scene flow via test-time optimization, can be used on any number of points, and show no dataset bias.}
    \begin{center}
    \scalebox{0.8}{
    \begin{tabular}{ccccrrrr}
			Dataset & Method & Supervision & Num. Points & EPE [m]~$\downarrow$ & $Acc_{.05}$ [\%]~$\uparrow$ & $Acc_{.10}$ [\%]~$\uparrow$ & $\theta_{\epsilon}$[$^\circ$]~$\downarrow$ \\
            \hline
			\multirow{7}{*}{Waymo} 
			& FLOT~\cite{puy2020flot} & \textcolor{RedOrange}{Full} & 8192 & 0.702 & 2.46 & 11.30 & 0.808\\
			& PointPWC-Net~\cite{wu2020pointpwc} & \textcolor{NavyBlue}{Self} & 8192 & 4.109 & 0.05 & 0.36 & 1.742 \\
			& WsRSF \cite{gojcic2021weakly} & \textcolor{RawSienna}{Weak} & 8192 & 0.414 & 35.47 & 44.96  & 0.527 \\ 
            & NSFP \cite{li2021nsfp} & \textcolor{ForestGreen}{None} & 8192 & 0.098 & 69.34 & 85.95 & 0.302 \\         
            & \textbf{MBNSF (Ours)} & \textcolor{ForestGreen}{None} & 8192 & \textbf{0.097} & \textbf{74.15} & \textbf{89.24}  & \textbf{0.295} \\
            \cline{2-8}
            & NSFP \cite{li2021nsfp} & \textcolor{ForestGreen}{None} & All & 0.091 & 76.61 & 89.00  & 0.282 \\
            & \textbf{MBNSF (Ours)} & \textcolor{ForestGreen}{None} & All & \textbf{0.066} & \textbf{82.29} & \textbf{92.44}  & \textbf{0.277} \\
            \hline
            \multirow{7}{*}{Argoverse} 
			& FLOT~\cite{puy2020flot} & \textcolor{RedOrange}{Full} & 8192 & 0.796 & 2.30 & 9.87 & 0.929 \\
			& PointPWC-Net~\cite{wu2020pointpwc} & \textcolor{NavyBlue}{Self} & 8192 & 5.724 & 0.02 & 0.15 & 1.147 \\
			& WsRSF \cite{gojcic2021weakly} & \textcolor{RawSienna}{Weak} & 8192 & 0.416 & 34.52 & 43.10  & 0.558 \\ 
            & NSFP \cite{li2021nsfp} & \textcolor{ForestGreen}{None} & 8192 & 0.056 & 70.30 & 89.57 & 0.265\\         
            & \textbf{MBNSF (Ours)} & \textcolor{ForestGreen}{None} & 8192 & \textbf{0.051} & \textbf{79.36} & \textbf{92.37}  & \textbf{0.264} \\
            \cline{2-8}
            & NSFP \cite{li2021nsfp} & \textcolor{ForestGreen}{None} & All & 0.090 & 64.97 & 83.09  & 0.230 \\            
            & \textbf{MBNSF (Ours)} & \textcolor{ForestGreen}{None} & All & \textbf{0.033} & \textbf{89.34} & \textbf{95.91}  & \textbf{0.168} \\
			\bottomrule
	\end{tabular}
	}
    \end{center} %
    \label{tab:scene_flow}
\end{table*}

We do not constrain the range of the point clouds and use the entire point cloud after ground-plane removal. 
We evaluate separately the learning-based methods that require pre-processing and down-sampling to 8192 points. Note that, unless otherwise specified, we do not compensate for ego-motion, and the pseudo ground truth scene flow contains the combination of both the absolute motion of scene elements and the relative motion caused by ego-motion. Our test set for Argoverse consists of 18 sequences with 25 consecutive frames for evaluating long-term trajectories (which also results in 450 pairs for scene flow evaluation), and our Waymo test set consists of 212 pairs for scene flow evaluation.

\paragraph{Evaluation Metrics}
For the scene flow evaluation in~\cref{sec:scene_flow_eval}, we employ the widely used metrics as in~\cite{liu2019flownet3d, wu2020pointpwc, li2021nsfp, najibi2022motion}, which are: 
\textit{3D End-Point-Error (EPE [m])}: the mean L2 distance between the flow prediction and ground truth for all points in the source point cloud. 
\textit{Strict Accuracy ($Acc_{.05}$ [\%])}: the percentage of points with \textit{EPE} $ < 0.05 \, m$ or relative error $< 5\%$.
\textit{Relaxed Accuracy ($Acc_{.10}$ [\%])}: the percentage of points with \textit{EPE} $ < 0.10 \, m$ or relative error $< 10\%$.
\textit{Angle Error ($\theta_{\epsilon}$ [$^\circ$])}: the mean angle error between the predicted flow vector and ground truth.
For long-term trajectory evaluation in~\cref{sec:traj_eval}, we follow the metrics in~\cite{wang2022ntp} and use $0.5 m$ and $1 m$ thresholds for strict (\textit{$Acc_{.5}$}) and relaxed (\textit{$Acc_{1.0}$}) accuracy, respectively, computed between the 1st and 25th frames in the sequence.

\subsection{Scene Flow Evaluation}
\label{sec:scene_flow_eval}

Evaluation of scene flow on the Argoverse and Waymo datasets is shown in~\cref{tab:scene_flow}. We compare methods under two settings: using only 8192 points (which is a prerequisite for all learning-based methods) and using all points in the raw point cloud (after ground-plane removal). We observe that learning-based methods drastically drop in performance in these large-scale real-world datasets, which are out of their training distribution, thus highlighting their main drawback. Test-time optimization methods (NSFP~\cite{li2021nsfp} and ours) are not susceptible to these drawbacks and can operate in any setting under any data distribution. We observe that our method MBNSF consistently and significantly improves the performance of NSFP across all metrics in these benchmark datasets. 

Additionally, we provide a quantitative comparison with NSFP++~\cite{najibi2022motion}. Since NSFP++ operates only on the subset of dynamic points after ego-motion compensation, we evaluate it separately. These results are included in the supplementary in~\cref{sec:nsfp_pp_results} and~\cref{tab:scene_flow_dynamic}.

\subsection{Trajectory Prediction Evaluation}
\label{sec:traj_eval}
\begin{table}[!t]
    \caption{Evaluation of long-term point-wise trajectory prediction using scene flow integration and direct 4D trajectory prediction. $^*$ indicates results presented in \cite{wang2022ntp}. $\dagger$ indicates our implementation.}
    \setlength{\tabcolsep}{4pt}
    \renewcommand{\arraystretch}{1.25}
    \begin{center}
	\resizebox{\columnwidth}{!}{
    \begin{tabular}{cccc}
			Approach & Method  & $Acc_{.5}$ [\%]~$\uparrow$ & $Acc_{1.0}$ [\%]~$\uparrow$  \\
            \hline
            \multirow{3}{*}{\begin{tabular}{c}\emph{Scene flow} \emph{integration}\\ \emph{(forward-Euler)} \end{tabular}} & NSFP~\cite{wang2022ntp}$^*$   & 45.28 & 59.52 \\
            & NSFP$\dagger$    & 38.09 & 54.20 \\
			& \textbf{MBNSF (Ours)}   & \textbf{61.11} & \textbf{74.34} \\
			\hline
			\multirow{3}{*}{\begin{tabular}{c}\emph{Direct 4D trajectory} \\ \emph{prediction} \end{tabular}}
			& NTP~\cite{wang2022ntp}$^*$    & 52.28 & 69.88 \\
            & NTP$\dagger$    & 48.06 & 62.45 \\
			& \textbf{MBNT (Ours)}   & \textbf{59.82} & \textbf{73.57} \\
	\end{tabular}
	}
    \end{center}
	\vspace{-0.3cm}
	\label{tab:traj_integration}
\end{table}

The evaluation of long-term trajectory prediction is presented in~\cref{tab:traj_integration}. We follow the same evaluation setup used in~\cite{wang2022ntp} and test our implementations of NSFP (with forward-Euler integration) and NTP against the results published in~\cite{wang2022ntp}. We also integrate our regularizer with NSFP and NTP. We observe that our regularizer can be used to significantly improve the performance of both NSFP and NTP and that our methods outperform the published results for both scene flow integration and direct 4D trajectory prediction, setting the new state-of-the-art for these tasks.

\subsection{Sensitivity to Cluster Size}
\label{sec:ablation}
\begin{figure}[t]
\centering
\includegraphics[width=0.8\linewidth]{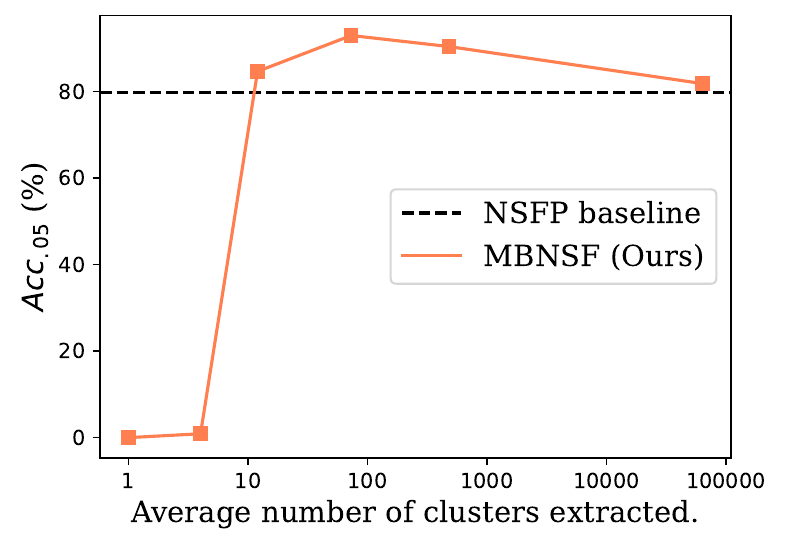}
\caption{Sensitivity to the size of clusters used to approximate rigid bodies. 
A large number of clusters corresponds to a small average cluster size, which leads to single-point clusters at the extremity (right). A small number of clusters implies large cluster sizes, which combine multiple rigid bodies into one cluster, resulting in the whole point cloud being a single cluster at the extremity (left).  The horizontal dashed line shows the baseline NSFP performance (without regularization). 
The x-axis is in log-scale. }
\label{fig:dbscan_param}
\end{figure}
The sensitivity to cluster size is shown in~\cref{fig:dbscan_param}. We vary the average size of the clusters extracted by varying the $\mathtt{DBSCAN}$ parameters \footnote{See Sec. \ref{sec:sup_cluster_size_sensitivity} for details on the corresponding $\mathtt{DBSCAN}$ parameters.}. When increasing the cluster size (\ie, decreasing the average number of clusters extracted), we first see an increase in performance (at optimal cluster size) and then a decrease as the regularizer becomes noisy (\eg, by combining multiple independently moving rigid bodies into one cluster). On the right extremity---as the number of clusters approaches the number of points---we see that the results converge to the baseline NSFP performance (it is still marginally higher as there are still some clusters with ${>}1$ point). This shows that our method is not dependent on the exact estimation of rigid bodies into one cluster, and that over-clustering (which may disrupt other methods) will gradually remove the multi-body rigidity constraint while retaining the accuracy of the baseline NSFP. 

On the left extremity---as the whole point cloud becomes one cluster---we see that the regularizer is now enforcing an incorrect objective (\eg, by combining multiple independently moving rigid bodies into one cluster), and the performance drops drastically. Although we would expect the lower margin of accuracy to correspond to the accuracy obtained by ego-motion compensation (applying a single transformation corresponding to ego-motion to the whole point cloud), our optimization does not give this result. We observe that as the size of the cluster grows larger, the regularizer tends to discourage motion, as lack of motion is guaranteed to preserve isometry. We note this as a limitation of our method. However, this limitation can be easily avoided by over-clustering, as discussed above. 

\begin{figure}[t]
     \centering
     \begin{subfigure}[b]{0.49\columnwidth}
         \centering
         \includegraphics[width=\linewidth]{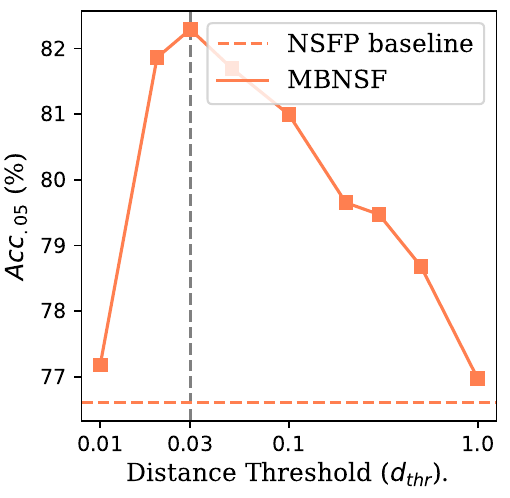}
         \caption{Accuracy}
         \label{fig:sup_dthr_ablation_accuracy}
     \end{subfigure}
     \hfill
     \begin{subfigure}[b]{0.49\columnwidth}
         \centering
         \includegraphics[width=\linewidth]{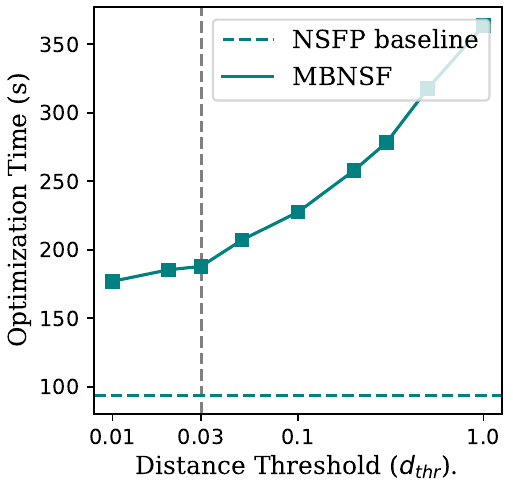}
         \caption{Optimization time}
         \label{fig:sup_dthr_ablation_efficiency}
     \end{subfigure}
        \caption{Variation of scene flow accuracy $Acc_{.05}$ (\cref{fig:sup_dthr_ablation_accuracy}) and optimization time (\cref{fig:sup_dthr_ablation_efficiency}) with variation in the distance threshold $d_{thr}$ (in Eq. \ref{eq:spatial_consistency}). We select the optimal value $d_{thr} = 0.03 m$ (vertical gray dashed line) for our experiments. The x-axis is in log-scale for both plots.}
        \vspace{-5mm}
        \label{fig:sup_dthresh_ablation}
\end{figure}

\begin{figure*}[t]
     \centering
     \begin{subfigure}[b]{0.33\textwidth}%
         \centering
         \includegraphics[width=\textwidth]{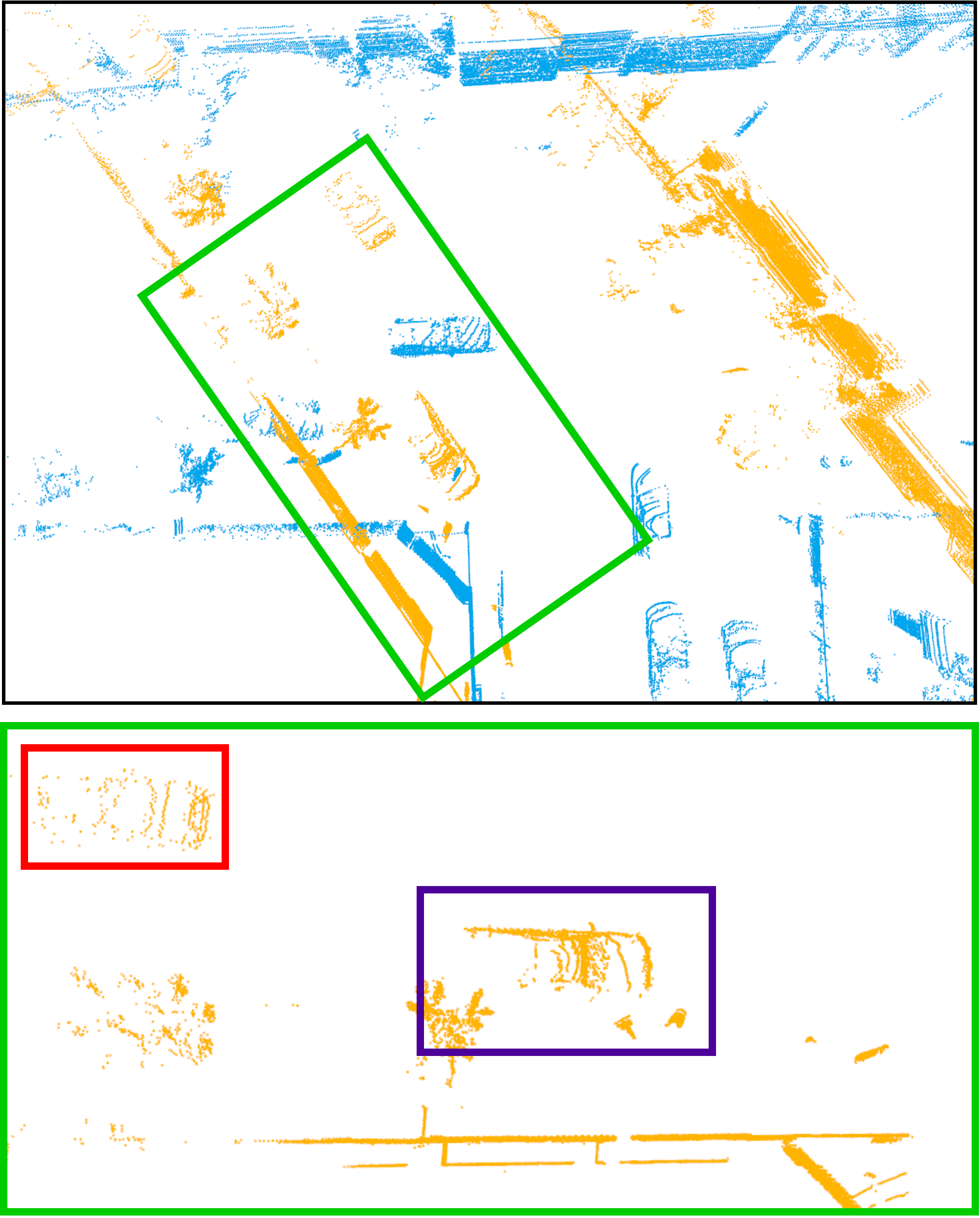}
         \caption{Before projection.}
         \label{fig:sc_viz_before}
     \end{subfigure}
     \hfill
     \begin{subfigure}[b]{0.33\textwidth}
         \centering
         \includegraphics[width=\textwidth]{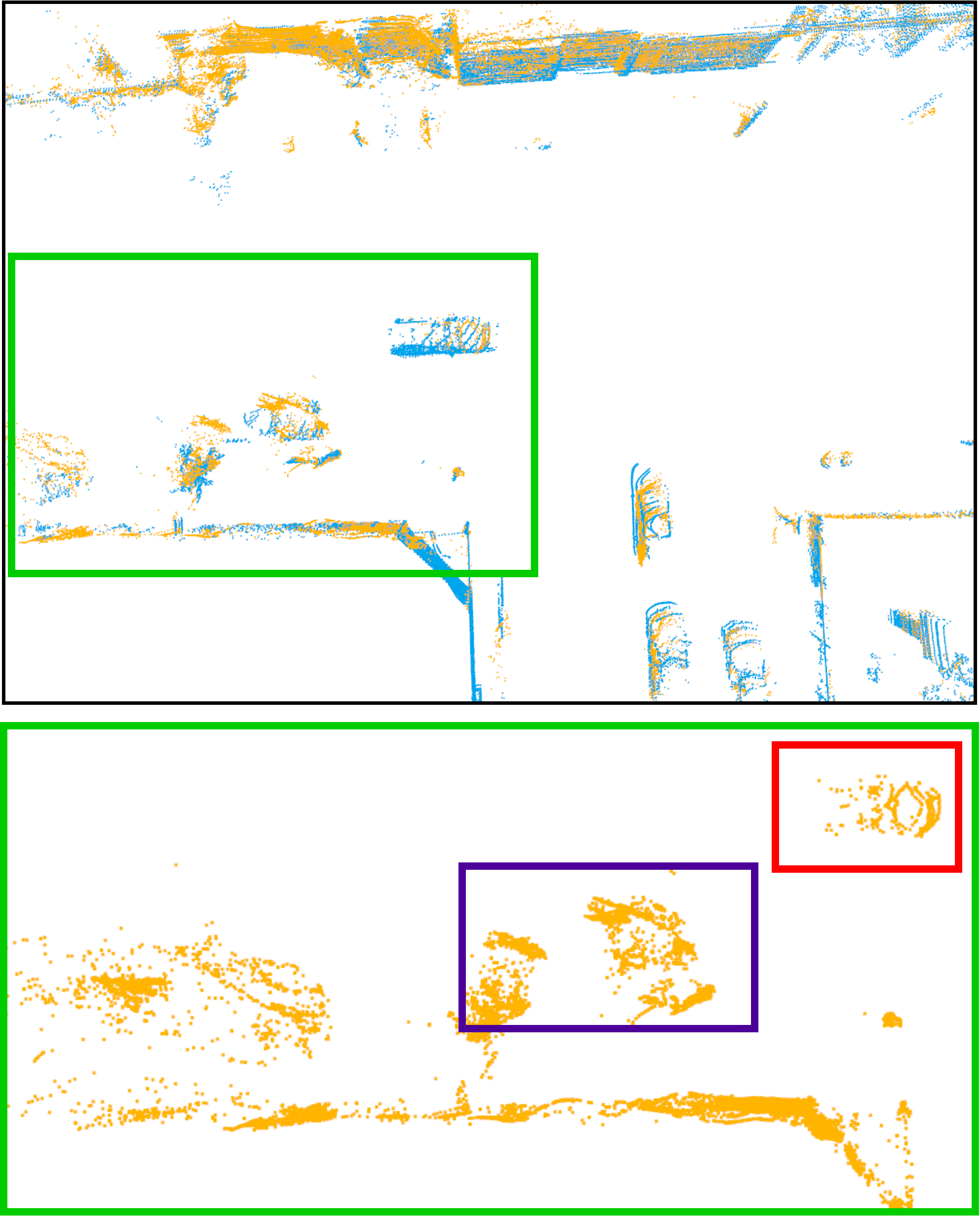}
         \caption{Project using NSFP.}
         \label{fig:sc_viz_nsfp}
     \end{subfigure}
     \hfill
     \begin{subfigure}[b]{0.33\textwidth}
         \centering
         \includegraphics[width=\textwidth]{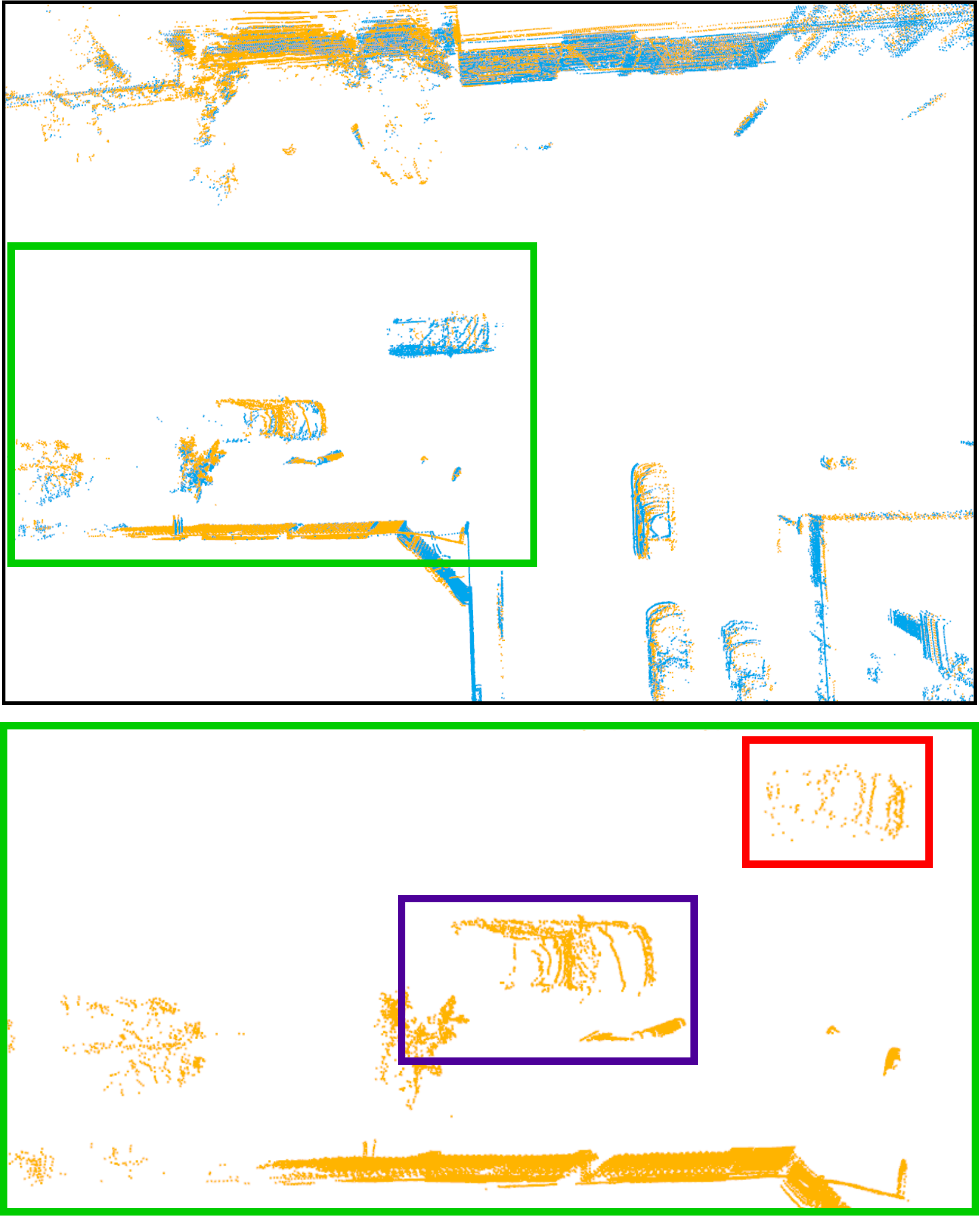}
         \caption{Project using MBNSF (Ours).}
         \label{fig:sc_viz_ours}
     \end{subfigure}
    \caption{Visualization of projecting the source cloud (yellow) to the target (blue), which is 25 frames ($2.5\,s$) apart, using forward Euler integration of scene flow. The second row is a zoom-in of the green box in the first row. Note the motion and shape of the purple and red cars - in this example, the red car overtakes the purple car when moving from source to target.
    \cref{fig:sc_viz_before}: there is a large motion between the source and target as the ego-vehicle turns at an intersection.
    \cref{fig:sc_viz_nsfp}: 
    NSFP has roughly aligned the motions of all points, but the shapes of rigid bodies are now deformed. Note how the purple car (in the second column) is deformed and no longer looks like a car.
    \cref{fig:sc_viz_ours}:
    MBNSF (Ours) has aligned the motions while preserving the shapes of all rigid bodies.
    }
    \vspace{-3mm}
    \label{fig:sc_viz}
\end{figure*}

\subsection{Sensitivity to the Distance Threshold ($d_{thr}$)}
\label{sec:sup_dthr_sensitivity}
We analyze the sensitivity to the distance threshold $d_{thr}$ (in~\cref{eq:spatial_consistency}) on the Waymo dataset~\cite{sun2020scalabilitywaymodataset} test set.
\cref{fig:sup_dthr_ablation_accuracy} shows the variation of scene flow accuracy with varying $d_{thr}$. We observe that increasing $d_{thr}$ initially increases and then decreases scene flow accuracy. Small values of $d_{thr}$ imply a strict margin on the approximate isometry enforced by our regularizer, and decreasing this value converges to an exact isometry (at the left extremity of the plot), which can be overly strict and impractical in real-world data containing noise. Conversely, increasing $d_{thr}$ to large values (at the right extremity of the plot) also gradually leads to lower performance as now the flow vectors are allowed to deviate by large margins, and rigidity is no longer enforced to a satisfactory extent. We observe that the right extremity naturally converges to the baseline NSFP performance (\ie, without regularization).
We select the optimal value of $d_{thr}\:{=}\:0.03 \, m$, which is a margin approximately equal to the noise present in modern LiDAR sensors~\cite{lidar2019velodyne}.

\cref{fig:sup_dthr_ablation_efficiency} shows the variation of average optimization time with varying distance threshold $d_{thr}$. We observe that the optimization time monotonically increases with $d_{thr}$.
Increasing $d_{thr}$ corresponds to increasing the size of the main cluster of the graph $\mathcal{G}$ approximated in~\cref{eq:sc_score_final} (as a less strict threshold allows more nodes to be compatible). Thus, increasing $d_{thr}$ corresponds to increasing the number of nodes (flow vectors) used in the optimization by our spatial regularizer, thereby increasing the optimization time.

\subsection{Qualitative Results}

Qualitative results depicting the enforcement of multi-body rigidity are depicted in~\cref{fig:sc_viz}. We present long-term scene flow integration results using forward-Euler integration for NSFP and MBNSF (Ours).
While the NSFP baseline is capable of producing visually accurate flow predictions for a single pair of point clouds, when estimating point trajectories across a long sequence of point clouds using the integration of pair-wise scene flow estimates, we start to notice the accumulation of error. By introducing our multi-body rigidity constraint, we can reduce these erroneous flow predictions and maintain multi-body rigidity even throughout long-term sequences.
Other multi-body methods, \eg NSFP++~\cite{najibi2022motion}, cannot handle such long-term motion, as discussed in~\cref{sec:nsfp_pp_results}.
Further qualitative results are in Sec. \ref{sec:sup_qualitative_results}.

\subsection{Limitations}
\label{sec:limitations}
\vspace{-1mm}
The incorporation of our regularizer decreases the efficiency of NSFP. On the Waymo dataset, under identical hyper-parameters, MBNSF roughly doubles the optimization time (from $93.8\,s$ to $186.3\,s$, as in~\cref{fig:sup_dthr_ablation_efficiency}). However, 
the main application of our method is in the offline setting where there are no real-time requirements. For example, NSFP (or our method) can be used to generate accurate labels, which can be used by any real-time supervised method, as in~\cite{vedder2023zeroflow}. %
Further, any subsequent speedups to NSFP (such as~\cite{li2023fast}) will also improve the efficiency of MBNSF. 
We also observe that the majority of the errors of MBNSF are located on dynamic objects, as discussed in Sec. \ref{sec:sup_qualitative_results_error_viz}.
Additionally, as discussed in Sec. \ref{sec:ablation}, under-clustering (combining multiple rigid bodies into one cluster) can cause our method to fail. However, this can be easily avoided by over-clustering.
\vspace{-0.3cm}

\section{Conclusion}
\label{sec:conclusion}

\vspace{-1mm}
The unsupervised estimation of scene flow enables many open-world 3D perception tasks.
To facilitate progress in this domain, we aim to improve the accuracy and robustness of neural prior-based test-time optimization of scene flow. 
We demonstrate how multi-body rigidity can be enforced without estimating the $SE(3)$ parameters of each rigid body.
This results in scene flow predictions that preserve multi-body rigidity while still maintaining a continuous flow field.  
We qualitatively demonstrate how our regularizer enforces multi-body rigidity in neural prior-based methods, thus eliminating their potential drawbacks of predicting physically implausible motion vectors. 
We further demonstrate that, unlike other methods, ours is not dependent on the accurate estimation of rigid bodies
as it is robust against over-clustering. 
The effects of these properties result in obtaining state-of-the-art performance for scene flow estimation and long-term point-wise trajectory prediction.%

\hfill \break
\noindent \textbf{Acknowledgement:} 
We thank Chaoyang Wang for sharing their implementation of NTP~\cite{wang2022ntp}. We thank Hemanth Saratchandran and Lachlan MacDonald for insightful discussions and technical feedback. 
{
    \small
    \bibliographystyle{ieeenat_fullname}
    \bibliography{main}
}
\clearpage
\setcounter{page}{1}
\maketitlesupplementary

\section{Overview}
In this supplementary document, we first provide more details on the implementation of our regularizer for enforcing multi-body rigidity in neural scene flow and long-term continuous motion fields (in~\cref{sec:sup_integration_details}).
We provide a tabular overview of related work and a qualitative and quantitative comparison with the most similar method to ours, \ie, NSFP++~\cite{najibi2022motion} (in~\cref{sec:sup_related_work}).
We analyze the sensitivity to the loss weight ($\omega$) in~\cref{sec:sup_omega_sensitivity}.
We provide additional details on the sensitivity to the cluster size (in~\cref{sec:sup_cluster_size_sensitivity}) to supplement the results in~\cref{sec:ablation} of the main paper. 
We provide additional qualitative results (in~\cref{sec:sup_qualitative_results}).

\section{Implementation Details}
\label{sec:sup_integration_details}
We provide details on how our regularizer can be integrated with NSFP~\cite{li2021nsfp} for enforcing multi-body rigidity into the continuous motion fields estimated by neural priors (\cref{sec:sup_integration_details_nsfp}), and also provide details on how to extend this concept into long-term trajectories using the formulation of neural trajectory prior NTP~\cite{wang2022ntp} (\cref{sec:sup_integration_details_ntp}). Additionally, we provide details on the power iteration algorithm used in~\cref{eq:sc_score_final} of the main paper (in~\cref{sec:sup_power_iteration}).

\subsection{Integration with NSFP (MBNSF)}
\label{sec:sup_integration_details_nsfp}

\begin{figure}[t]
\centering
\includegraphics[width=0.99\linewidth]{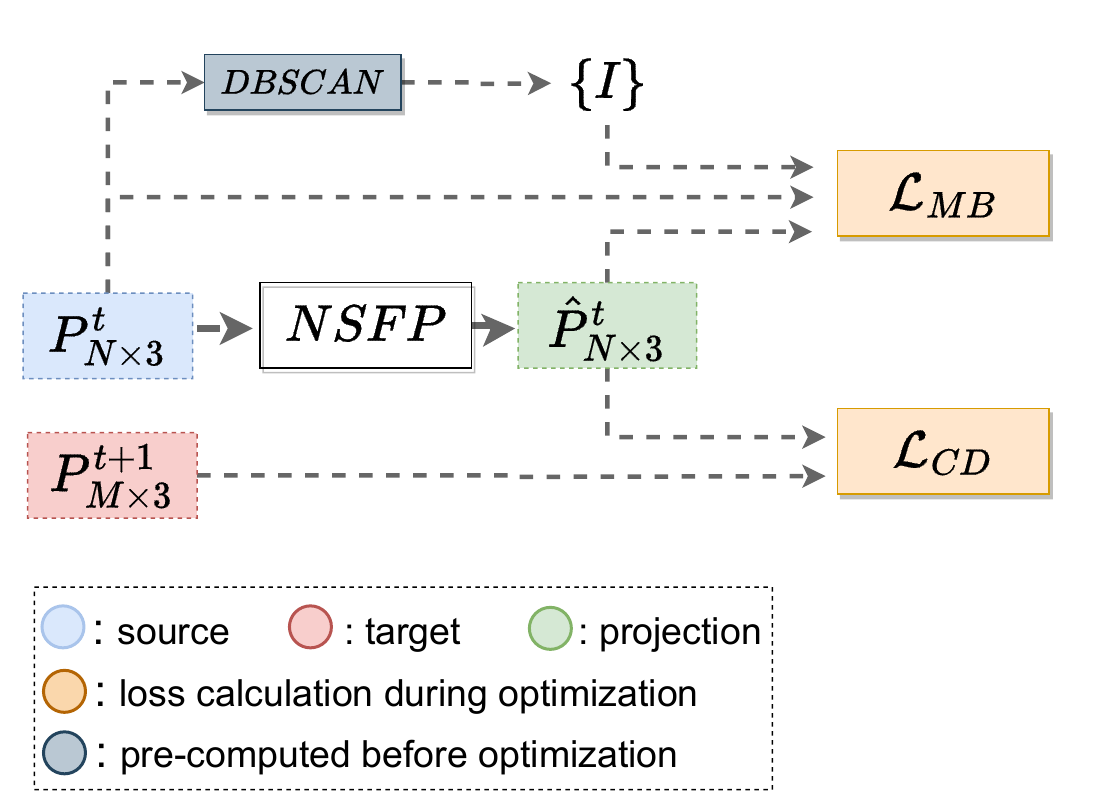}
\caption{MBNSF - Integration of our multi-body regularizer with NSFP~\cite{li2021nsfp}. $\mathcal{L}_{CD}$ is the truncated Chamfer distance loss utilized in the original NSFP. $\mathcal{L}_{MB}$ is our proposed regularization term (\ie the output of~\cref{alg:method} in the main paper). $\{\mathbf{I}\}$ represents the cluster indices extracted using $\mathtt{DBSCAN}$ \cite{ester1996dbscan}. 
}
\label{fig:sup_nsfp_integration}
\end{figure}

\cref{fig:sup_nsfp_integration} depicts a block diagram of how our multi-body regularizer can be integrated with the optimization of NSFP~\cite{li2021nsfp}. The source point cloud $ \mathbf{P}^{t} \:{\in}\: \mathbb{R}^{N {\times} 3}$ at time $t$ (blue) is projected (green) to match the target point cloud $ \mathbf{P}^{t+1} \:{\in}\: \mathbb{R}^{M {\times} 3}$ at time $t{+}1$ (red).

We highlight two key properties of our formulation and why they are advantageous:
\begin{itemize}
    \item $\mathtt{DBSCAN}$ is only needed on the source cloud, and this clustering process is only performed once (\ie, before the start of the optimization iterations). This formulation is more efficient than the alternatives. 
    \item Our regularizer loss $\mathcal{L}_{MB}$ is independent of the target cloud (red), \ie, our method enforces multi-body rigidity without the need for the cumbersome and brittle process of estimating frame-to-frame cluster correspondences as in NSFP++~\cite{najibi2022motion}.
\end{itemize}

\subsection{Integration with NTP (MBNT)}
\label{sec:sup_integration_details_ntp}
\begin{figure*}[t]
\centering
\includegraphics[width=0.8\linewidth]{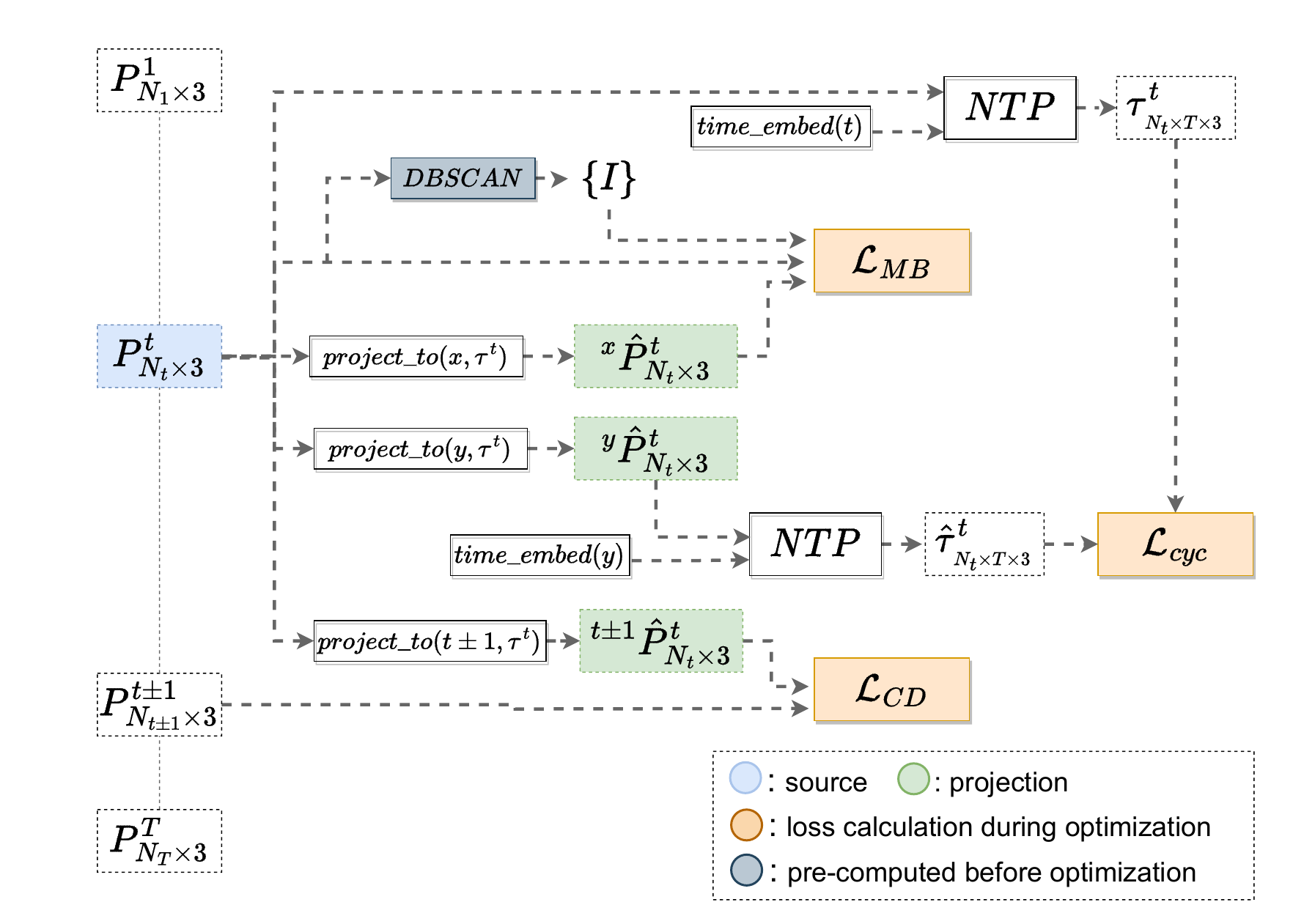}
\caption{MBNT - Integration of our multi-body regularizer with NTP~\cite{wang2022ntp} for estimating the long-term motion field of the sequence of $T$ point cloud frames $ \{\mathbf{P}^i\}_{i=1}^{T}$. $\mathcal{L}_{CD}$ is the truncated Chamfer distance loss, and $\mathcal{L}_{cyc}$ is the cycle-consistency loss utilized in the original NTP. $\mathcal{L}_{MB}$ is our proposed regularization (\ie, the output of~\cref{alg:method} in the main paper). x,y are arbitrary time stamps in [1,T]. $\mathcal{L}_{CD}$ is only applied to $t \pm 1$ as the Chamfer distance cannot handle long-term changes~\cite{wang2022ntp}. $\mathcal{L}_{cyc}$ and $\mathcal{L}_{MB}$ can be applied to any/all timestamps in [1,T]. The function $\mathtt{time\_embed(\cdot)}$ represents the cosine positional encoding utilized in NTP. The function $\mathtt{project\_to(x, \tau)}$ represents projecting the input point cloud to the timestamp $x$ using the trajectory output $\tau$ of NTP.
}
\label{fig:sup_ntp_integration}
\end{figure*}

\cref{fig:sup_ntp_integration} depicts a block diagram of how our multi-body regularizer can be integrated with the optimization of NTP~\cite{wang2022ntp} to estimate long-term continuous motion fields. This diagram depicts a single training step for optimizing the parameters of the NTP architecture---in each training step, the source point cloud (blue) is selected at random from the sequence $ \{\mathbf{P}^i\}_{i=1}^{T}$. We note that the $\mathtt{DBSCAN}$ clustering process is carried out for each point cloud in the sequence only once, and this is done before the start of the optimization process. We additionally note that the complexity of this diagram arises from the complexity of NTP and that the integration of our regularizer to NTP remains simple and identical to its integration to NSFP in~\cref{fig:sup_nsfp_integration}.

This highlights that our formulation for incorporating multi-body rigidity in neural-prior-based scene flow and trajectory estimation is simple and elegant. Yet, it does not compromise any of the advantageous properties of neural priors (such as the estimation of continuous motion fields or the fully unsupervised nature). 

\subsection{Power Iteration Algorithm}
\label{sec:sup_power_iteration}
 The power iteration algorithm can be used to efficiently approximate the leading eigenvector $\mathbf{v}^*$ of a matrix $\mathbf{A}$ in an iterative manner as 
 \begin{equation}
    \mathbf{v}^{*}_{k+1} = \frac{\mathbf{A}\mathbf{v}^{*}_{k}}{\lVert\mathbf{A}\mathbf{v}^{*}_{k} \rVert}, \quad \mathbf{v}^{*}_{0} = \mathbf{1}
 \end{equation}
where $k$ represents the iteration index. We initialize $\mathbf{v}^{*}_{0}$ as a vector of ones and use 10 iterations to approximate $\mathbf{v}^*$ used in~\cref{eq:sc_score_final} of the main paper.

\section{Comparison with Related Work}
\label{sec:sup_related_work}
\cref{tab:sup_related_work} shows a tabular overview of related works that propose spatial regularization of scene flow on LiDAR data. Our method is the only one that can enforce multi-body rigidity while:
\begin{itemize}
    \item Maintaining a continuous motion field.
    \item Not directly constraining the $SE(3)$ parameters of each rigid body. 
    \item Supporting the estimation of long-term trajectories. 
\end{itemize}
\begin{table*}[!t]
    \caption{Tabular overview of published literary works that incorporate spatial regularization of scene flow on LiDAR data. Methods are sorted chronologically (in terms of publication date). Bold values represent the preferred/advantageous option. }
    \begin{center}
    \resizebox{2\columnwidth}{!}{
    \begin{threeparttable}[t]
    \begin{tabular}{ccccccccc}
            \toprule
			\multirow{2}{*}{Method} & Multi-Body & Label & Test-Time & Full & Long-Term & Continuous & Open\\
            & Rigidity~\tnote{1} & Independence & Optimization & $SE(3)$~\tnote{2}  & Trajectory~\tnote{3} & Motion Field & Source \\
            \hline
            Graph-Laplacian~\cite{pontes2020graphlaplacian} & - & \cmark & \cmark & - & - & - & - \\
            NSFP~\cite{li2021nsfp}~\tnote{4} & - & \cmark & \cmark & - & - & \cmark & \cmark \\
            WsRSF~\cite{gojcic2021weakly} & \cmark ($\mathtt{DBSCAN}$ clusters) & - & - & \cmark (direct) & - & - & \cmark \\
            ExploitingRigidity~\cite{dong2022exploiting} & \cmark ($\mathtt{DBSCAN}$ clusters) & - & - & \cmark (direct) & - & - & \cmark \\
            RigidFlow~\cite{li2022rigidflow} & \cmark (super-voxels) & \cmark & - & \cmark (direct) & - & - & - \\
            NTP~\cite{wang2022ntp}~\tnote{4} & - & \cmark & \cmark & - & \cmark & \cmark & - \\
            Huang \etal~\cite{huang2022dynamic} & \cmark ($\mathtt{DBSCAN}$ clusters) & - & - & \cmark (direct) & - & - & \cmark \\
            NSFP++~\cite{najibi2022motion} & \cmark ($\mathtt{DBSCAN}$ clusters) & - & \cmark & - & - & - & - \\
            RSF~\cite{deng2023rsf} & \cmark (bounding-boxes) & \cmark & \cmark & \cmark (direct) & - & - & - \\
            \textbf{Ours} & \cmark ($\mathtt{DBSCAN}$ clusters) & \cmark & \cmark & \cmark (\textbf{indirect}) & \cmark & \cmark & \cmark \\
			\bottomrule
	\end{tabular}
 \begin{tablenotes}
    \item[1] `Multi-Body Rigidity' denotes the explicit awareness of the multiple rigid bodies in the scene during spatial regularization. 
    \item[2] `Full $SE(3)$' denotes constraining both the translation and rotation components of motion for rigid bodies. 
    \item[3] `Long-Term Trajectory' denotes estimating long-term trajectories of points given a sequence of point clouds (beyond 2 frames). 
     \item[4] NSFP~\cite{li2021nsfp} and NTP~\cite{wang2022ntp} are implicit spatial regularizers, while all other methods are explicit.
   \end{tablenotes}
   \end{threeparttable}
   }
    \end{center} %
    \label{tab:sup_related_work}
\end{table*}

We note that:
\begin{itemize}
    \item RSF~\cite{deng2023rsf} is the only other method that enforces multi-body rigidity in a fully unsupervised manner in a test-time optimization setting. 
    \item NSFP++~\cite{najibi2022motion} is the only other method that enforces multi-body rigidity using neural priors.
\end{itemize}

\subsection{Comparison with NSFP++}
\label{sec:nsfp_pp_results}

\paragraph{Technical Comparison:}We list five key differences in our method compared to NSFP++~\cite{najibi2022motion} and why they are advantageous. 
Our method:
(1) Only requires optimizing only one NSPF network per scene, whereas NSFP++ optimizes one per cluster. We note that optimizing an NSPF network per cluster is resource-intensive and encourages increasing the average cluster size (to reduce the number of clusters extracted per scene), which we show can lead to a noisy regularizer in~\cref{sec:ablation}.
(2) Enforces full $SE(3)$ rigidity, whereas NSFP++ only constrains the translation component, which makes it susceptible to predicting physically implausible flow predictions as the rotation is unaccounted for.
(3) Is fully unsupervised, whereas NSFP++ is dependent on ego-motion compensation.
(4) Predicts a continuous motion field for the scene (NSFP++ is only continuous within each cluster).
(5) Can easily be extended into long-term sequences (as in~\cref{sec:traj_eval}), whereas NSFP++ relies on cluster correspondence estimation via a ``box-query expansion'' strategy that accumulates errors across multiple frames. 

\begin{table}[!t]
    \caption{Evaluation of scene flow on dynamic points. Input point clouds are ego-motion compensated and filtered to extract only dynamic points. $\dagger$ indicates our implementation.}
    \setlength{\tabcolsep}{4pt}
    \renewcommand{\arraystretch}{1.25}
    \begin{center}
	\resizebox{\columnwidth}{!}{
    \begin{tabular}{cccccc}
			Method & EPE [m]~$\downarrow$  & $Acc_{.05}$ [\%]~$\uparrow$ & $Acc_{.10}$ [\%]~$\uparrow$ & $\theta_{\epsilon}$[$^\circ$]~$\downarrow$ & Time [s] \\
            \hline
            NSFP\cite{li2021nsfp} & 0.463   & 19.08 & 39.91 & 0.575 & 8.02 \\
            NSFP++\cite{najibi2022motion}$\dagger$    & \textbf{0.228 }& \underline{39.33} & \textbf{71.59} & \underline{0.340} & 37.43 \\
			MBNSF (Ours)   & \underline{0.237} & \textbf{44.56} & \underline{71.46} & \textbf{0.285} & 24.36 \\

	\end{tabular}
	}
    \end{center}
	\label{tab:scene_flow_dynamic}
\end{table}

\paragraph{Quantitative Comparison:}Additionally, we quantitatively compare our method with NSFP++~\cite{najibi2022motion}. Since NSFP++ operates only on dynamic point clouds after ego-motion compensation, we prepare a separate test split using our Waymo test split, which is pre-processed to include ego-motion compensation and static point filtering. The results under this setting are shown in~\cref{tab:scene_flow_dynamic} for baseline NSFP~\cite{li2021nsfp} using the original implementation, NSFP++~\cite{najibi2022motion} using our implementation, and NSFP with our regularizer (denoted as MBNSF). 

We observe that both NSFP++ and our method demonstrate improvements in performance. The technical drawback of NSFP++ only constraining the translation component of rigid motion is not adequately represented in available datasets as the majority of the data consists of the ego-vehicle and other dynamic elements moving on straight roads without rotation. 

\paragraph{Memory Complexity Comparison:}Although both NSFP++ and our method obtain comparable performance, we note that the memory complexity of NSFP++ is $\mathcal{O}(mN)$, where $N$ is the number of parameters of the MLP and $m$ is the number of clusters. 
Our memory complexity is $\mathcal{O}(N)$ since we only optimize $1$ MLP per scene. In large scenes with many dynamic elements, $m$ can be in the order of $100$s, potentially limiting the practicality of NSFP++.

\paragraph{Efficiency Comparison:}Additionally, we note that NSFP++ has lower efficiency due to the need for optimizing a separate NSFP network on each cluster, whereas our method only requires optimizing one network per scene. In the scenarios represented in this test set (which only consists of dynamic elements after filtering static points), we see that this results in NSFP++ taking $53.65\%$ longer than ours to optimize ($37.43\,s$ vs. $24.36\,s$), on average. We note that if the whole point cloud were used, this time gap would be much larger as additional clusters would be present. 

\section{Sensitivity to the loss weight ($\omega$)}
\label{sec:sup_omega_sensitivity}
\begin{figure}[t]
     \centering
     \begin{subfigure}[b]{0.49\columnwidth}
         \centering
         \includegraphics[width=\linewidth]{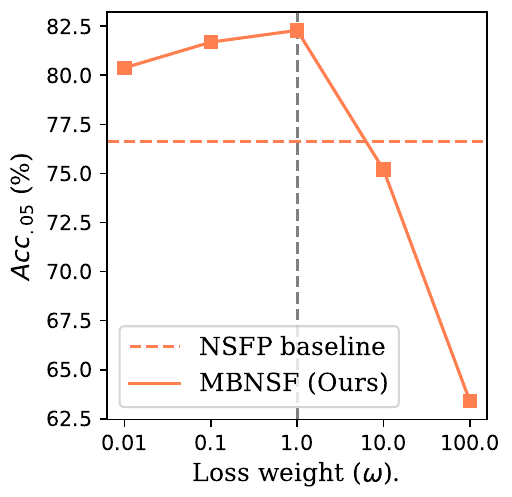}
         \caption{Accuracy}
         \label{fig:sup_lw_ablation_accuracy}
     \end{subfigure}
     \hfill
     \begin{subfigure}[b]{0.49\columnwidth}
         \centering
         \includegraphics[width=\linewidth]{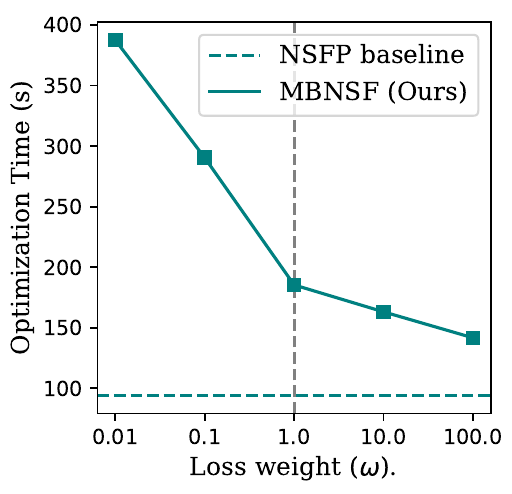}
         \caption{Optimization time}
         \label{fig:sup_lw_ablation_efficiency}
     \end{subfigure}
        \caption{Variation of scene flow accuracy $Acc_{.05}$ (\cref{fig:sup_lw_ablation_accuracy}) and optimization time (\cref{fig:sup_lw_ablation_efficiency}) with variation in the loss weight $\omega$  (in Eq. \ref{eq:final_objective}). We select the optimal value $\omega = 1.0$ (vertical gray dashed line) for our experiments. The x-axis is in log-scale for both plots.}
        \label{fig:sup_lw_ablation}
\end{figure}

In this section, we discuss the sensitivity to the loss weight $\omega$ (in~\cref{eq:final_objective}) on the Waymo dataset~\cite{sun2020scalabilitywaymodataset} test set.
\cref{fig:sup_lw_ablation_accuracy} shows the variation of scene flow accuracy with varying loss weight $\omega$. We observe that increasing $\omega$ initially leads to an increase in accuracy (as the effect of our regularizer increases) up to a certain extent and then leads to a decrease. 
While the left extremity naturally converges to the baseline NSFP performance (\ie, without regularisation), the right extremity drops below the baseline performance. This is due to higher values of $\omega$ over-powering the $\mathcal{L}_{CD}$ Chamfer distance loss, which is solely responsible for aligning the source cloud to the target. Higher values of $\omega$ may discourage point-wise motion as lack of motion guarantees maintaining isometry, similar to the behavior noted in~\cref{sec:limitations}.

We observe in~\cref{fig:sup_lw_ablation_efficiency} that increasing $\omega$ consistently leads to a decrease in optimization time. This is due to lower values of $\omega$ requiring more iterations to optimize for multi-body rigidity.

\section{Sensitivity to Cluster Size}
\label{sec:sup_cluster_size_sensitivity}
In~\cref{sec:sup_dthr_sensitivity} of the main paper, we demonstrated the sensitivity of the accuracy of our method to the average size of the clusters extracted. In this section, we provide supplementary details to those results by analyzing the sensitivity to efficiency (in \cref{sec:sup_cluster_size_sensitivity_efficiency}) and visualizing the clusters for the utilized data points (in \cref{sec:sup_cluster_size_sensitivity_qualitative}).

\subsection{Efficiency}
\label{sec:sup_cluster_size_sensitivity_efficiency}
We analyze how varying the average cluster size affects the runtime of our method. \cref{tab:sup_dbablation} shows the runtime values related to the 6 data points used in~\cref{fig:dbscan_param} of the main paper. This study was done on a single sequence of the Argoverse test set. We gradually increase the average cluster size by varying the 2 $\mathtt{DBSCAN}$~\cite{ester1996dbscan} parameters:`$eps$' (aka Epsilon, the radius used to check density around each point), and `$min\_points$' (the minimum number of points needed to form a cluster). We note that as the average cluster size increases (descending rows), the average number of clusters extracted (column 3) decreases. We use the open-source implementation of $\mathtt{DBSCAN}$ from the Open3D library~\cite{zhou2018open3d}.

We compare the runtime of 2 components: 
$T_{C}$ represents the time taken by $\mathtt{DBSCAN}$ to extract clusters. $T_{\mathcal{L}}$ represents the time taken by~\cref{alg:method} (in the main paper) to calculate $\mathcal{L}_{MB}$. We note that these 2 runtimes are not additive since $T_{C}$ is only incurred  once for each point cloud pair, but $\mathcal{L}_{MB}$ is incurred at each optimization iteration. 

We observe that the time for extracting clusters ($T_{C}$) increases with the average cluster size. However, this runtime is only incurred once (during pre-processing) per each point cloud pair and therefore is not a significant portion of the total runtime. The runtime for the calculation of $\mathcal{L}_{MB}$ ($T_{\mathcal{L}}$) increases with increasing number of clusters. This is due to the score for each cluster (\cref{eq:sc_score_final} in the main paper) being calculated in a \textit{for-loop}, as in~\cref{alg:method}. However, we note that this increase is sub-linear since as the number of clusters increases, the average size of clusters decreases; hence the compute time for each cluster decreases. 

\begin{table}%
    \caption{Evaluation of the effect of average cluster size on the runtime. $T_{C}$ represents the time taken by $\mathtt{DBSCAN}$ to extract clusters. $T_{\mathcal{L}}$ represents the time taken (by the \textit{for-loop} in~\cref{alg:method}) to calculate $\mathcal{L}_{MB}$. }
    \setlength{\tabcolsep}{4pt}
    \renewcommand{\arraystretch}{1.25}
    \begin{center}
	\resizebox{\columnwidth}{!}{
    \begin{tabular}{rr|r|crr}
			\multicolumn{2}{c|}{$\mathtt{DBSCAN}$ parameters} & Avg. Clusters & \multirow{2}{*}{ $Acc_{.05}$ [\%]~$\uparrow$ } & \multirow{2}{*}{$T_{C}$ [s]} & \multirow{2}{*}{$T_{\mathcal{L}}$ [s]} \\
            \cline{1-2}
            $eps$ [m] & $min\_points$ & Extracted & & \\
            \hline
            0.01 & 1   & 63432.5 & 81.80 & 0.08 & 1637.5 \\
            0.10   & 6 & 481.5 & 90.35 & 0.11 & 333.6 \\
			0.80   & 30 & 73.0 & \textbf{92.91} & 1.57 & 47.3 \\
            5.00 & 100   & 12.0 & 84.61 & 19.70 & 63.3 \\
            10.00   & 200 & 4.0 & 00.89 & 32.40 & 48.2 \\
			15.00   & 300 & 1.0 & 00.00 & 40.07 & 50.9\\

	\end{tabular}
	}
    \end{center}
	\label{tab:sup_dbablation}
\end{table}

From the results in~\cref{tab:sup_dbablation}, we see that our runtime increases sub-linearly with the number of clusters. Additionally, as demonstrated in~\cref{sec:sup_dthr_sensitivity} of the main paper, our formulation for enforcing multi-body rigidity does not depend on optimal extraction of each rigid body into one cluster, and through gradual over-clustering (\ie, splitting single rigid bodies into multiple clusters), our method naturally converges to the baseline performance (without regularization). In other words, over-clustering does not hinder the accuracy of our method, and its runtime increases sub-linearly with the number of clusters. We use the parameters $eps \:{=}\: 0.8 \, m$ and $min\_points \:{=}\: 30$ for generating results
in the main paper.

\subsection{Cluster Visualization}
\label{sec:sup_cluster_size_sensitivity_qualitative}
\begin{figure*}[t]
\centering
\includegraphics[width=0.99\linewidth]{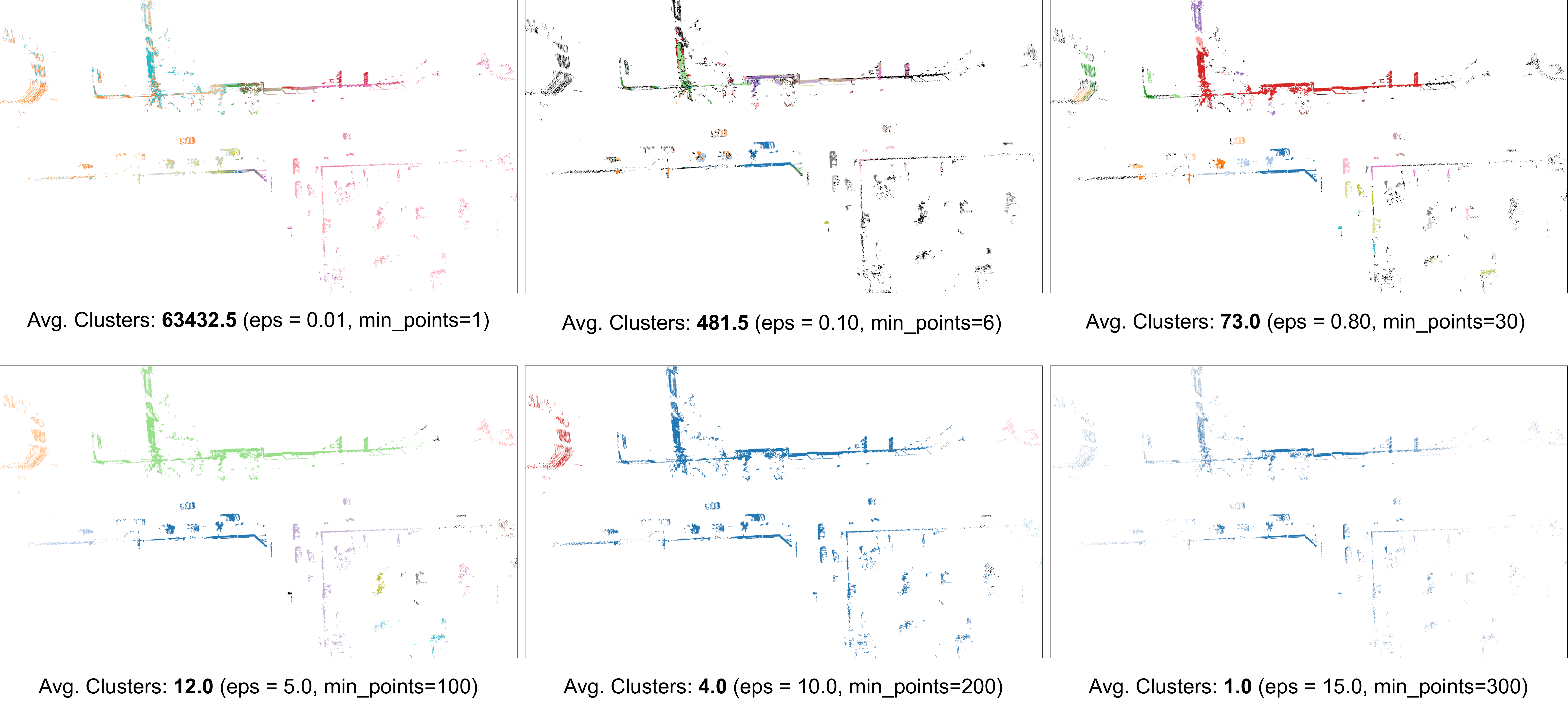}
\caption{Visualization of extracting rigid bodies using various cluster sizes. This figure depicts the cluster visualization of the 6 data points used to create \cref{fig:dbscan_param} in the main paper.  We vary the $\mathtt{DBSCAN}$ parameters to control the average size of the clusters extracted. Note that as the average size of the clusters increases, the average number of clusters extracted per scene decreases, and vice-versa.  The sub-captions indicate the average number of clusters extracted for the 2 $\mathtt{DBSCAN}$ parameters given in brackets. Point clouds have been pre-processed to remove the ground-plane and are visualized from a top-down orthogonal point-of-view. Extracted clusters are colored randomly, with each cluster represented by a unique color. Black-colored points are points that have not been assigned to any cluster, and our regularizer does not act on these points. 
}
\label{fig:sup_cluster_size_datapoint_viz}
\end{figure*}
\cref{fig:sup_cluster_size_datapoint_viz} shows a visualization of the clusters extracted at various sizes (on the Argoverse dataset \cite{chang2019argoverse} test set) related to the 6 data points used in~\cref{fig:dbscan_param} of the main paper. As the average size of clusters increases (left-to-right and top-to-bottom), the average number of clusters extracted decreases, resulting in the whole point cloud becoming one cluster at the extremity.

\section{Qualitative Results}
\label{sec:sup_qualitative_results}

In this section, we provide additional qualitative results of our method. \cref{sec:sup_qualitative_results_SC} presents visual results on the enforcement of multi-body rigidity in scene flow via our regularizer. \cref{sec:sup_qualitative_results_error_viz} presents visual results on the spatial distribution of the end-point-error. 

\subsection{Multi-Body Rigidity}
\label{sec:sup_qualitative_results_SC}

\begin{figure*}[t]
     \centering
     \begin{subfigure}[b]{0.3\textwidth}
         \centering
         \includegraphics[width=\textwidth]{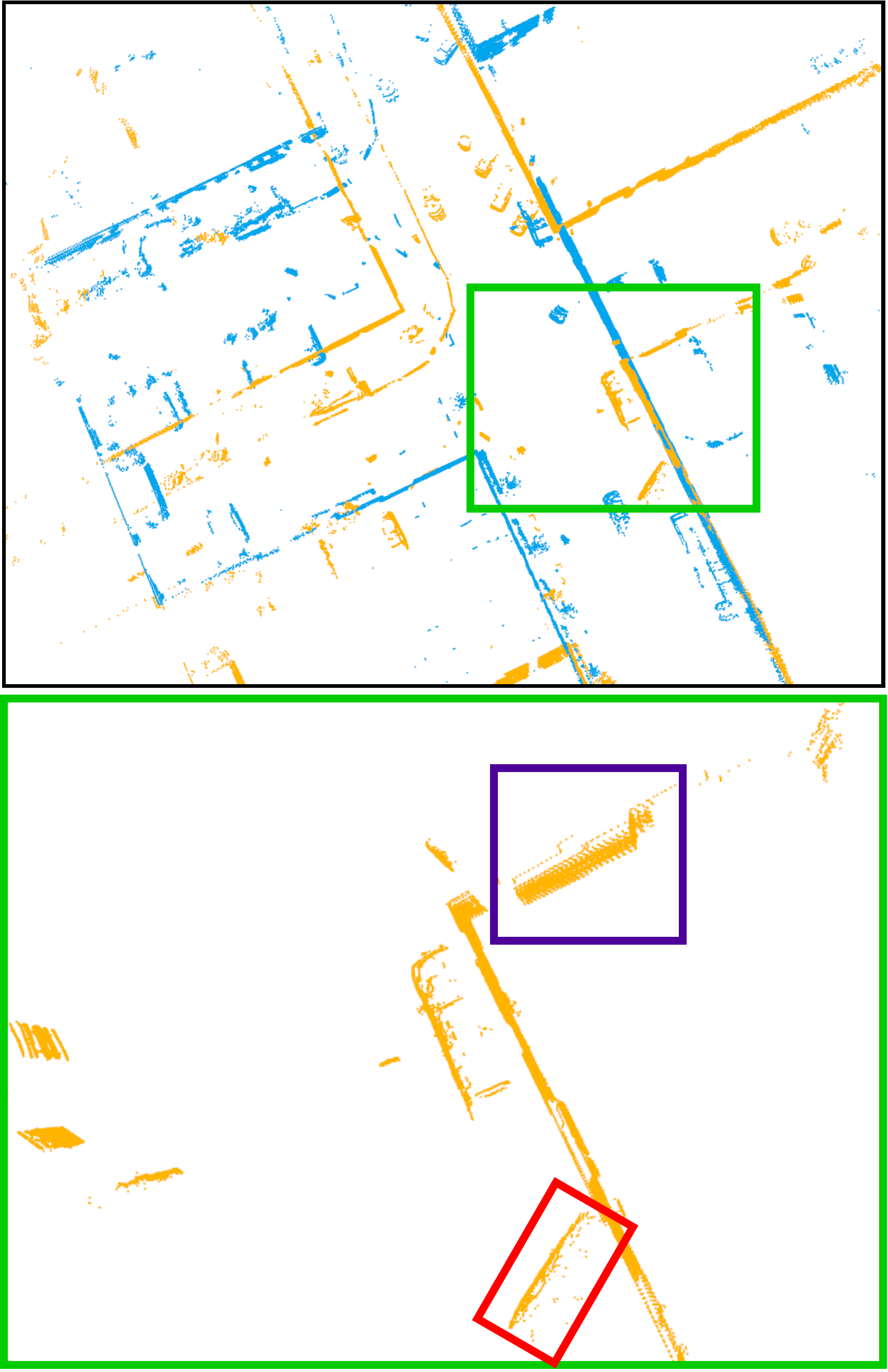}
         \caption{Before projection.}
         \label{fig:sup_sc_viz_before}
     \end{subfigure}
     \hfill
     \begin{subfigure}[b]{0.3\textwidth}
         \centering
         \includegraphics[width=\textwidth]{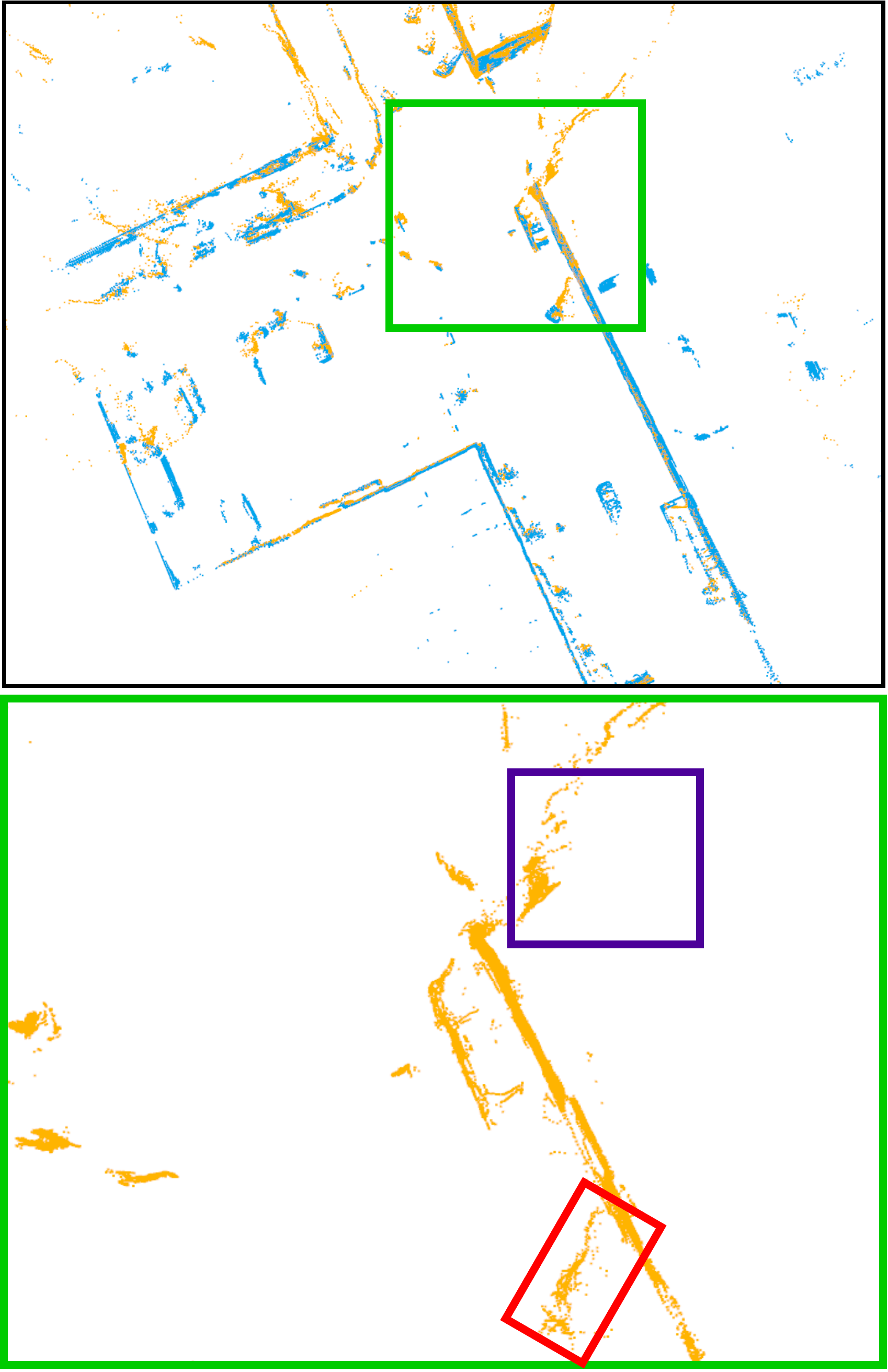}
         \caption{Project using NSFP.}
         \label{fig:sup_sc_viz_nsfp}
     \end{subfigure}
     \hfill
     \begin{subfigure}[b]{0.305\textwidth}
         \centering
         \includegraphics[width=\textwidth]{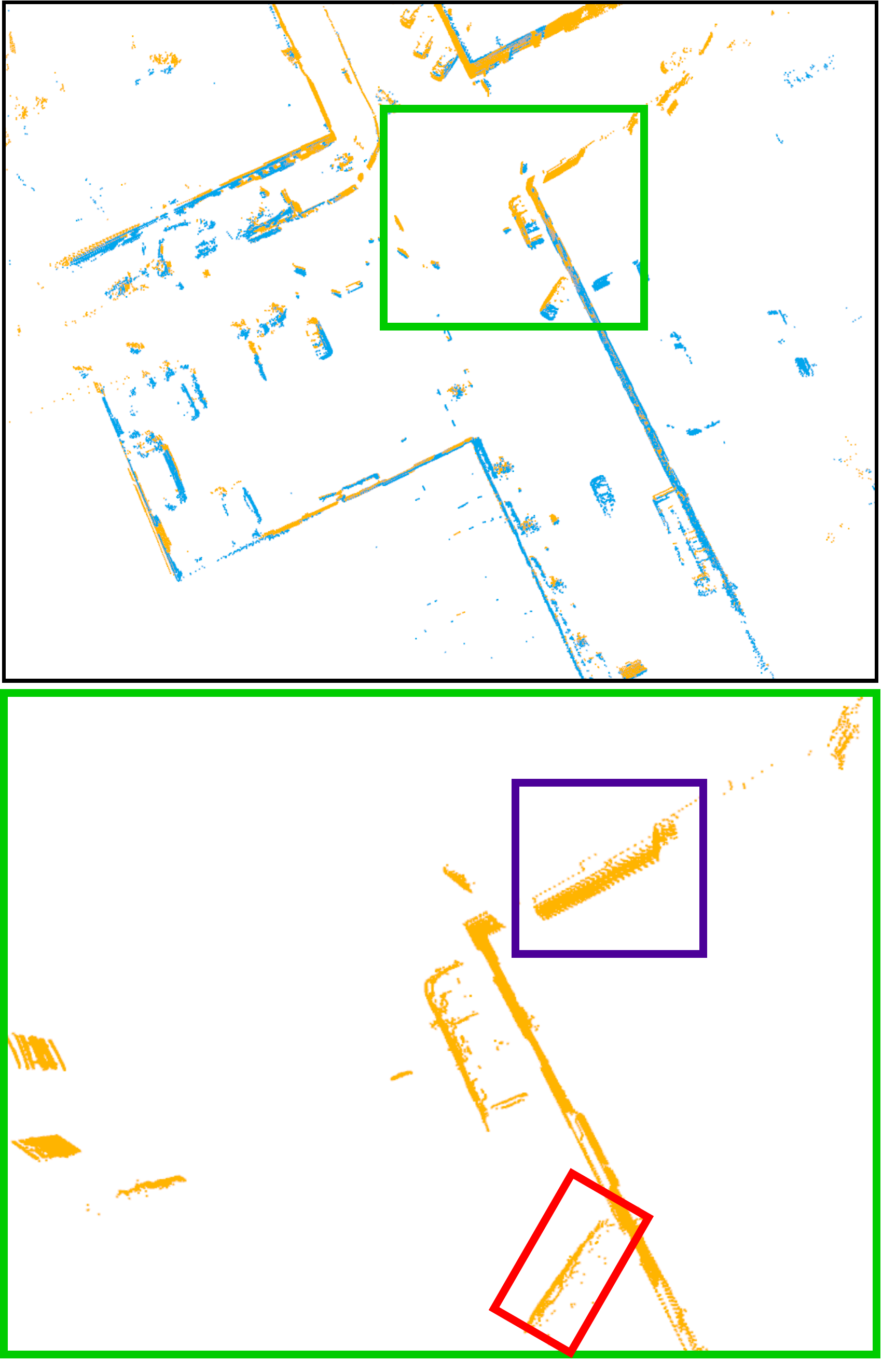}
         \caption{Project using MBNSF (Ours).}
         \label{fig:sup_sc_viz_ours}
     \end{subfigure}
    \caption{Visualization of projecting the source (yellow) to the target (blue), which is 25 frames (2.5 s) apart, using forward Euler integration. The second row is a zoom-in of the green box in the first row. 
    \cref{fig:sup_sc_viz_nsfp}: 
    NSFP has roughly aligned the motions of all points, but the shapes of rigid bodies are now deformed. Note how the purple wall (in the second column) is largely deformed.
    \cref{fig:sup_sc_viz_ours}:
    MBNSF (Ours) has aligned the motions while preserving the shapes of all rigid bodies.
    }
    \label{fig:sup_sc_viz}
\end{figure*}

Additional qualitative results depicting the enforcement of multi-body rigidity in flow predictions are depicted in \cref{fig:sup_sc_viz} to supplement the results in~\cref{fig:sc_viz} of the main paper. We present long-term scene flow integration results using NSFP in~\cref{fig:sup_sc_viz_nsfp} and MBNSF (Ours) in~\cref{fig:sup_sc_viz_ours}.
We again observe that while the NSFP baseline has as noticeably deformed the point cloud while minimizing the Chamfer distance, our method maintains the geometric structure of all rigid bodies and accurately predicts motion. 

\subsection{Error Visualization}
\label{sec:sup_qualitative_results_error_viz}

\begin{figure*}[t]
     \centering
     \begin{subfigure}[t]{0.3\textwidth}
         \centering
         \includegraphics[width=\textwidth]{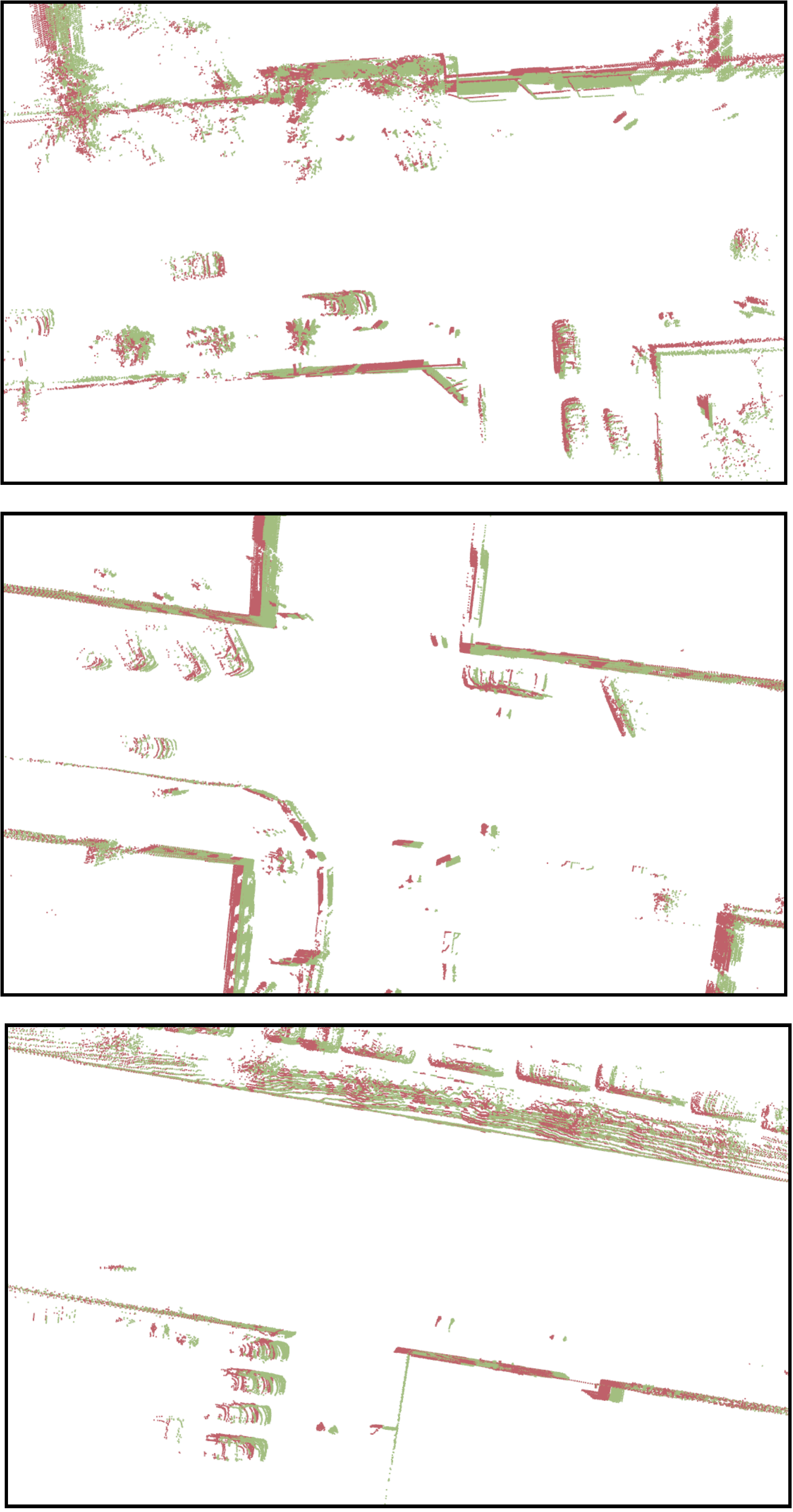}
         \caption{Source (green) and Target (red).}
         \label{fig:sup_error_viz_before}
     \end{subfigure}
     \hfill
     \begin{subfigure}[t]{0.3\textwidth}
         \centering
         \includegraphics[width=\textwidth]{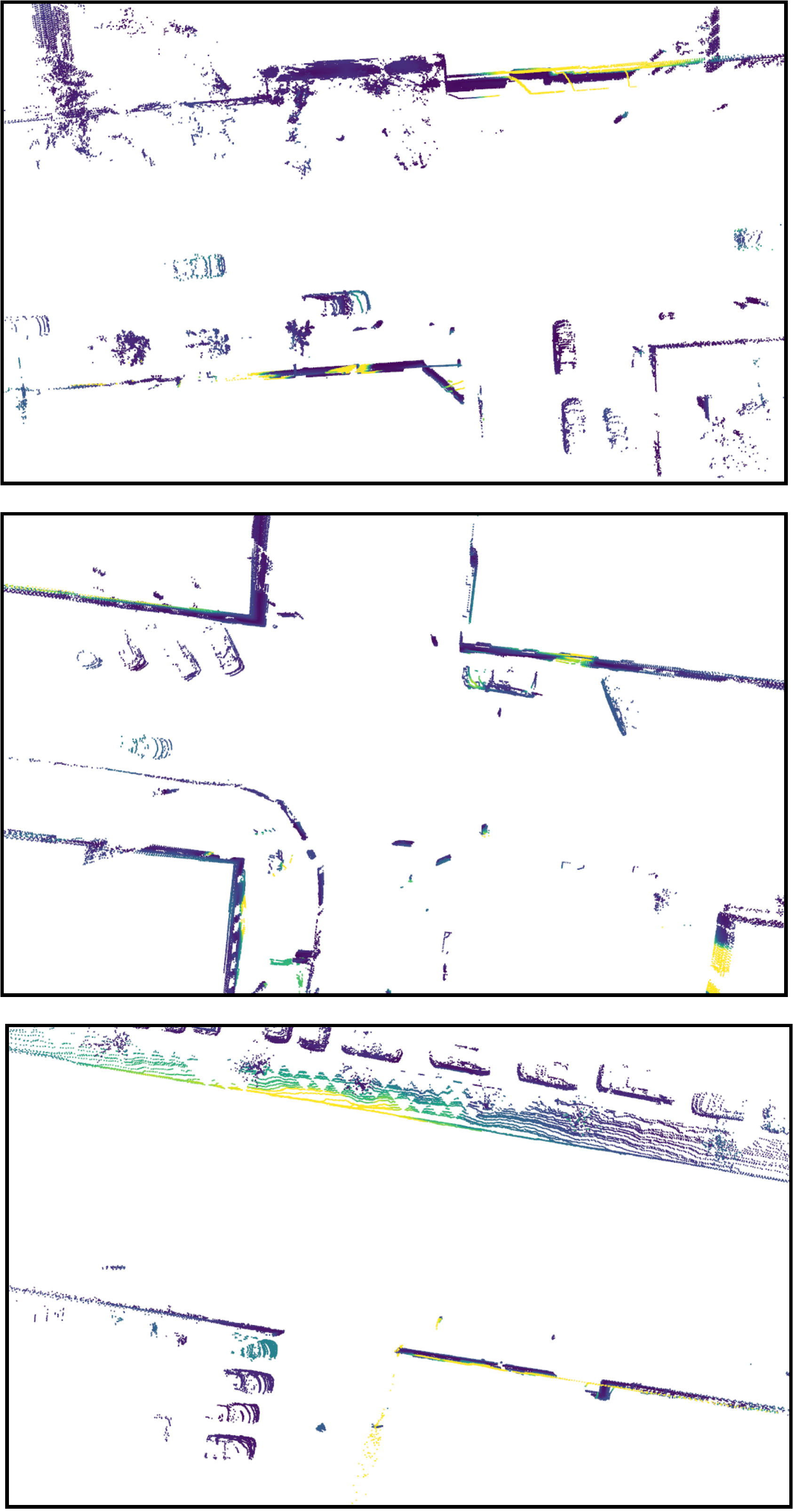}
         \caption{EPE for NSFP.}
         \label{fig:sup_error_viz_nsfp}
     \end{subfigure}
     \hfill
     \begin{subfigure}[t]{0.3\textwidth}
         \centering
         \includegraphics[width=\textwidth]{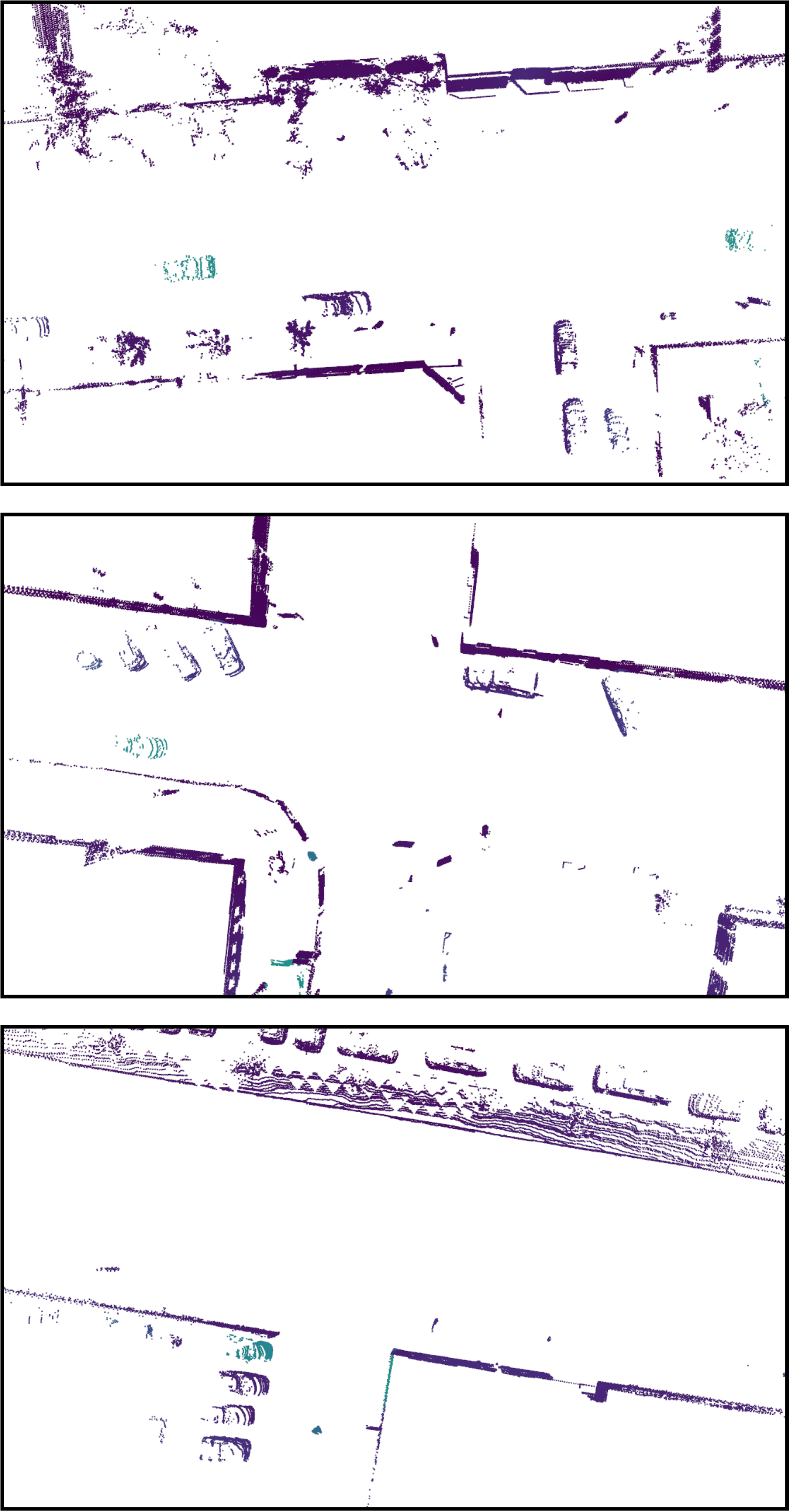}
         \caption{EPE for MBNSF (Ours).}
         \label{fig:sup_error_viz_ours}
     \end{subfigure}
     \begin{subfigure}[t]{0.05\textwidth}
         \centering
         \includegraphics[width=\textwidth]{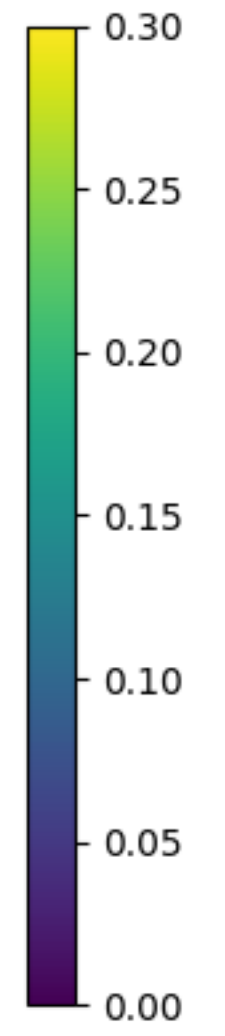}
     \end{subfigure}
    \caption{Visualization of End-Point-Error (EPE). The error color bar is shown on the right. 
    }
    \label{fig:sup_error_viz}
\end{figure*}

Visualization of end-point-error is shown in \cref{fig:sup_error_viz}. We visually analyze the spatial distribution of end-point-error on the Argoverse dataset for NSFP~\cite{li2021nsfp} and MBNSF (Ours). Errors are clipped at $ 0.3 \, m$ for visualization purposes.

Due to only minimizing the Chamfer distance during optimization, NSPF struggles to preserve geometric structure in areas that are inconsistently sampled by the LiDAR sensor between the source and the target frames. In other words, minimizing the Chamfer distance will try to find exact correspondence between points in the source and target frames when in reality, such correspondences do not exist. 

The incorporation of our regularizer mitigates this drawback by explicitly enforcing rigidity while minimizing Chamfer distance. We see in \cref{fig:sup_error_viz_ours} that our method has corrected the deviations of NSFP and that our method shows near zero end-point-error for all background and static elements in the scene. 

\cref{fig:sup_error_viz_ours} also highlights a limitation in our method. We see that most of the errors of MBNSF are on dynamic objects. We postulate that, for dynamic objects, there is a 'tug-of-war' between the Chamfer distance minimization and the multi-body loss minimization. This is because the trivial solution to minimizing the multi-body loss is to predict zero scene flow (since zero flow maintains isometry) or scene flow corresponding to the ego-motion (which is reinforced in the neural prior by the majority of background points). Further investigation of this interplay between the two loss terms is left for future work.


\end{document}